%% file: acl2023.tex
\pdfoutput=1

\documentclass[11pt]{article}

\usepackage{ACL2023}

\usepackage{times}
\usepackage{latexsym}
\usepackage[T1]{fontenc}
\usepackage[utf8]{inputenc}
\usepackage{microtype}
\usepackage{inconsolata}
\usepackage{graphicx}
\usepackage{booktabs}
\usepackage{multirow}
\usepackage{array}
\usepackage{hyperref}
\usepackage{enumitem}

\usepackage{tabularx}
\usepackage{xcolor}
\usepackage{colortbl}
\usepackage{caption}
\usepackage{microtype}
\usepackage{helvet}
\usepackage{amsmath}

\usepackage[normalem]{ulem}
\usepackage{soul}

\usepackage{tikz}
\usepackage{pgfplots}

\usepackage{dblfloatfix}

\usepackage{fontawesome5}   

\pgfplotsset{compat=1.18}
\usepgfplotslibrary{polar}
\usetikzlibrary{calc}

\definecolor{eventcol}{RGB}{50,90,160}
\definecolor{agencycol}{RGB}{40,130,70}
\definecolor{settingcol}{RGB}{180,120,20}
\definecolor{gutcol}{RGB}{55,138,221}
\definecolor{redcol}{RGB}{29,158,117}
\definecolor{wikicol}{RGB}{186,117,23}

\definecolor{eventbg}{RGB}{220, 235, 252}
\definecolor{agencybg}{RGB}{220, 250, 230}
\definecolor{settingbg}{RGB}{255, 243, 220}
\definecolor{eventheader}{RGB}{50, 90, 160}
\definecolor{agencyheader}{RGB}{40, 130, 70}
\definecolor{settingheader}{RGB}{180, 120, 20}
\definecolor{defbg}{RGB}{245, 245, 248}

\definecolor{agencycolor}{HTML}{9B8AC4}
\definecolor{settingcolor}{HTML}{5BC8A0}
\definecolor{eventcolor}{HTML}{6BAED6}

\definecolor{teagancolor}{RGB}{42,128,135}
\definecolor{maapcolor}{RGB}{194,100,50}
\definecolor{apcolor}{RGB}{210, 50, 50}

\definecolor{eventcolor}{HTML}{6BAED6}

\definecolor{focalcolor}{HTML}{6B5AA1}
\definecolor{emotioncolor}{HTML}{9B8AC4}
\definecolor{changecolor}{HTML}{C4B8DF}

\definecolor{concretecolor}{HTML}{3A9E74}
\definecolor{sensorycolor}{HTML}{8EDCB8}

\usepackage{xspace}
\usepackage{accsupp}   

\newcommand{\wordmark}[3][2.0ex]{%
  \BeginAccSupp{method=escape,ActualText={#3}}%
  \raisebox{-0.25ex}{\includegraphics[height=#1]{#2}}%
  \EndAccSupp{}\xspace}

\newcommand{\narradolma}{\wordmark{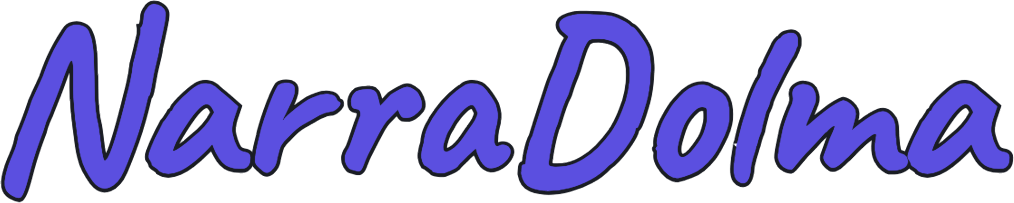}{NarraDolma}}
\newcommand{\narrabert}{\wordmark{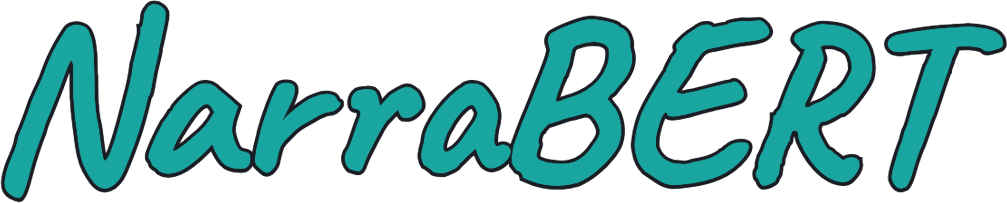}{NarraBERT}}

\DeclareRobustCommand{\wordmark}[3][2.0ex]{%
  \BeginAccSupp{method=escape,ActualText={#3}}%
  \raisebox{-0.25ex}{\includegraphics[height=#1]{#2}}%
  \EndAccSupp{}\xspace}

\DeclareRobustCommand{\narradolma}{\wordmark{images/logos/NarraDolma-wordmark.png}{NarraDolma}}
\DeclareRobustCommand{\narrabert}{\wordmark{images/logos/NarraBERT-wordmark.png}{NarraBERT}}

\input{commands}

\newcommand{\ulsolid}[2]{\textcolor{#2}{\uline{\textcolor{black}{#1}}}}
\newcommand{\uldash}[2]{\textcolor{#2}{\dashuline{\textcolor{black}{#1}}}}
\newcommand{\uldot}[2]{\textcolor{#2}{\dotuline{\textcolor{black}{#1}}}}

\newlength{\legkeywidth}
\newcommand{\legsolid}[1]{\raisebox{\ULdepth}{\textcolor{#1}{\uline{\makebox[\legkeywidth]{}}}}}
\newcommand{\legdash}[1]{\raisebox{\ULdepth}{\textcolor{#1}{\dashuline{\makebox[\legkeywidth]{}}}}}
\newcommand{\legdot}[1]{\raisebox{\ULdepth}{\textcolor{#1}{\dotuline{\makebox[\legkeywidth]{}}}}}

\newcommand{\score}[2]{%
  #1~\raisebox{0.15ex}{%
    \begin{tikzpicture}[baseline=0ex]
      \foreach \i in {1,...,5} {
        \ifnum\i>#1
          \fill[gray!25] (\i*0.26-0.24, 0) rectangle (\i*0.26, 0.16);
        \else
          \fill[#2] (\i*0.26-0.24, 0) rectangle (\i*0.26, 0.16);
        \fi
      }
    \end{tikzpicture}%
  }%
}

\newcommand{\scoresmall}[2]{%
  #1~\raisebox{0.15ex}{%
    \begin{tikzpicture}[baseline=0ex]
      \foreach \i in {1,...,5} {
        \ifnum\i>#1
          \fill[gray!25] (\i*0.18-0.16, 0) rectangle (\i*0.18, 0.12);
        \else
          \fill[#2] (\i*0.18-0.16, 0) rectangle (\i*0.18, 0.12);
        \fi
      }
    \end{tikzpicture}%
  }%
}

\newcommand{\propbar}[2]{%
  #1~\raisebox{0.15ex}{%
    \begin{tikzpicture}[baseline=0ex]
      \fill[gray!25] (0,0) rectangle (0.9, 0.12);
      \fill[#2] (0,0) rectangle (#1*0.9, 0.12);
    \end{tikzpicture}%
  }%
}

\newcommand{\hficon}{\raisebox{-0.8ex}{\includegraphics[height=3.4ex]{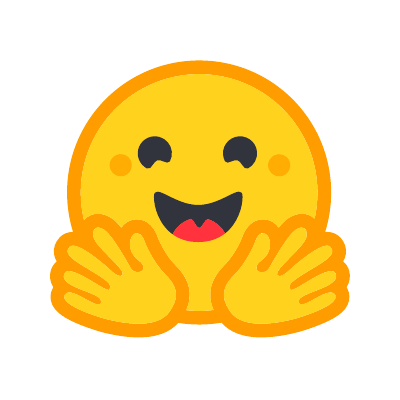}}}

\newcommand{\iconfootnote}[2]{{\renewcommand{\thefootnote}{#1}\footnote{#2}}}

\newcolumntype{L}{>{\raggedright\arraybackslash}X}
\newcolumntype{P}[1]{>{\raggedright\arraybackslash}p{#1}}

\title{Characterizing Narrative Content in Web-scale LLM Pretraining Data}

\newcommand{\affsym}[1]{\textsuperscript{\ensuremath{#1}}}

\author{
  \textbf{Teagan Johnson}\affsym{\clubsuit} \quad
  \textbf{Elliott Ash}\affsym{\spadesuit} \quad
  \textbf{Andrew Piper}\affsym{\heartsuit} \quad
  \textbf{Maria Antoniak}\affsym{\clubsuit} \\[0.6em]
  \small
  \affsym{\clubsuit}University of Colorado Boulder \quad
  \affsym{\spadesuit}ETH Z\"{u}rich \quad
  \affsym{\heartsuit}McGill University
}

\begin{document}
\maketitle

\begin{abstract}
The narrative composition of web-scale LLM pretraining corpora remains largely unexplored, even though narrative is a fundamental mode of human communication. We present the first fine-grained study of narrative features in \textsc{Dolma}, a 3-trillion-token open pretraining corpus. Drawing on narrative theory, we design a framework spanning three core narrative elements (agency, setting, and events) operationalized as 11 interpretable dimensions. 
After curating and hand-annotating a diverse set of $\sim$400 passages, we create an LLM-labeled dataset of 25K passages, and finally, we finetune and validate \textsc{NarraBERT}, two \textsc{RoBERTa}-based models for fine-grained narrative prediction.
We apply \textsc{NarraBERT} to $\sim$13M passages, resulting in a new dataset, \textsc{NarraDolma}. 
We find that narrative structure is measurable at scale across extremely heterogeneous data and narrative qualities are unequally distributed across pretraining sources, topics, and formats in ways that current data curation practices neither measure nor account for. Our framework, dataset, and analyses provide a foundation for understanding how narrative qualities are distributed in LLM pretraining data.
We publicly release \textsc{NarraDolma} and \textsc{NarraBERT}.\iconfootnote{\hficon}{\url{https://huggingface.co/collections/teagrjohnson/narratives-in-llm-pretraining-data}}\iconfootnote{\faGithub}{\url{https://github.com/johnsont4/narratives_in_pretraining_data_release}}
\end{abstract}

\section{Introduction}
\label{sec:intro}

Narratives are among the most pervasive and cognitively central forms of human communication. From ancient oral traditions to contemporary social media, people use narrative to record experience, build shared knowledge, and reason about cause and effect \cite{herman2011basic, boyd2009origin, gottschall2012storytelling}. 
Long a central subject of study in NLP~\citep{riedl2010narrative},
narrative tasks such as story generation have recently become one of the most popular use cases for large language models (LLMs)~\cite{gupta2026ai}. 

\begin{figure}[t]
    \centering
    \small
    \setlength{\fboxsep}{6pt}
    \fbox{%
    \begin{minipage}{0.90\columnwidth}
    \parbox[t]{\linewidth}{\raggedright\itshape
    ``I walked into the interview room and immediately 
    noticed the panel was twice the size I expected. My 
    hands started shaking. I sat down, took a breath, and 
    somehow got through the first question.''
    }\\[6pt]
    \renewcommand{\arraystretch}{1.15}
    \scriptsize
    \begin{tabular}{@{} l c @{\hspace{12pt}} l c @{}}
    \textbf{\textit{Agency}} & & \textbf{\textit{Setting}} & \\
    Focal. & \mbox{\score{4}{agencycolor}} & Concrete. & \mbox{\score{3}{settingcolor}} \\
    Emotion & \mbox{\score{4}{agencycolor}} & Temporal & \mbox{\score{1}{settingcolor}} \\
    Cognition & \mbox{\score{3}{agencycolor}} & Spatial & \mbox{\score{2}{settingcolor}} \\
    Change & \mbox{\score{3}{agencycolor}} & Sensory & \mbox{\score{2}{settingcolor}} \\
    Conflict & \mbox{\score{3}{agencycolor}} & & \\
    \end{tabular}\\[4pt]
    \begin{tabular}{@{} l c @{\hspace{12pt}} l c @{}}
    \textbf{\textit{Events}} & & & \\
    Temporal & \mbox{\propbar{1.00}{eventcolor}} & Causal & \mbox{\propbar{0.50}{eventcolor}} \\
    \end{tabular}\\[3pt]
    {\scriptsize Event Density: $\sim$\textbf{0.15} events per token}
    \end{minipage}%
    }
    \caption{An example web passage scored by \narrabert across our 12 narrative dimensions. Agency and setting dimensions are rated on a 1--5 Likert scale, temporal sequencing and causal density are passage-level proportions (0--1), and event density is the rate of event triggers per token. This passage scores high on agency and event features but low on setting, a \textbf{narrative profile} common to first-person web narratives.}
    \label{fig:teaser}
\end{figure}

The narrative capabilities of LLMs are shaped,
in part, by the narrative content present in their pretraining data. 
However, while recent work has examined pretraining data along dimensions such as quality, toxicity, deduplication, and topic distribution \cite{lucy-etal-2024-aboutme,wettig2025organize}, the narrative composition of these corpora has received almost no systematic attention. We do not know how much narrative content is present, how it is distributed across sub-corpora and genres, or how narrative-relevant features vary across the heterogeneous web text that dominates these datasets.
 

This gap matters for several reasons. If narrative content is unevenly concentrated in certain sub-corpora (e.g., books or Reddit), training mixtures that downweight those sources may disproportionately reduce a model's narrative exposure. Conversely, certain genres of narrative may be overrepresented in ways that skew models toward particular event structures, perspectives, narrative forms, or safety violations. The well-documented creativity deficit in LLM-generated storytelling may be related to training mixtures and not only preference tuning~\citep{tian2024large, chakrabarty2024art}.


In a first thorough investigation of narrative qualities of pretraining data, we map the narrative qualities of \textsc{Dolma}~\cite{dolma2024}, a 3-trillion-token open corpus spanning twelve sub-corpora. We build on the theoretical frameworks of \citet{herman2011basic} and \citet{piper-etal-2021-narrative} and define narrative as the structured sequencing of events through an agent's perspective in a grounded setting as told by a narrator. Given the dataset's extremely heterogeneous nature, \textbf{our work treats narrative as a continuous, multi-dimensional structure}~\citep{ochs2009living}, not as a binary classification \cite{pianzola2018looking,antoniak-etal-2024-people}. Our framework annotates passages along 11 narrative dimensions organized around three core elements from narrative theory: \textbf{events} (temporal ordering and causality), \textbf{agency} (focalization, emotion, cognition, change of state, and conflict), and \textbf{setting} (concreteness, temporal grounding, spatial grounding, and sensory detail). From a validated event detector, we additionally derive \textbf{event density} (events per token), giving the 12 features shown in Fig.~\ref{fig:teaser}.

Our contributions are as follows: 
\begin{itemize}
    \item an annotation framework grounded in narrative theory that operationalizes 11 dimensions across events, agency, and setting,
    \item \narrabert, a pair of efficient classifiers that predict all 11 narrative dimensions, validated against both human and LLM annotations,
    \item \narradolma, a dataset of 13M labeled passages sampled from the \textsc{Dolma} corpus using a principled stratified pipeline, and
    \item a large-scale analysis showing that narrativity in pretraining data forms a continuous, multidimensional structure and is unequally distributed across pretraining data categories.
\end{itemize}

A recurring finding across our analysis is that pretraining sources should not be treated as narratively homogeneous. 
This has direct implications for data curation: decisions made at the source level (e.g., which corpora to include, at what weight) are too coarse to account for the narrative diversity present in pretraining data. 

\section{Related Work}
\label{sec:related}

\subsection{Narrative Detection}
The focus on predicting the narrativity of texts or passages has gained traction in NLP in recent years, especially for social media content. For example, \citet{ganti-etal-2022-narrative} examine narrative detection in online health forums, and 
\citet{doyle2024stories} looks at narrative detection within the context of suicide bereavement forums on Reddit. 
These studies perform classification within the context of specific topics and online communities.

Two new datasets have been developed to support the task of narrative detection across topics. \textsc{StorySeeker} \cite{antoniak-etal-2024-people} provides span-level binary annotations on passages from Reddit for their narrative content. \textsc{NarraDetect} \cite{piper-bagga-2025-narradetect} provides a large binary-labeled dataset of passages drawn from 18 different genres across books and online texts for narrative content, as well as a small, manually annotated corpus for narrativity using a five-point Likert scale across agency, eventfulness, and world-building. 

We extend this work by annotating 11 fine-grained narrative dimensions on a 5-point Likert scale to provide a richer characterization of narrativity that is still computationally tractable.

\subsection{Web-Scale Text Corpora and Data Curation}
A growing body of work studies the composition of pretraining data~\citep{lucy-etal-2024-aboutme}. Data mixing research has shown that the source proportions of pretraining corpora influence performance on a range of benchmarks \cite{xie2023doremi, rae2021gopher, liu2025regmixdatamixtureregression, wettig2025organize}. RegMix \cite{liu2025regmixdatamixtureregression} learns optimal data mixing weights by training small models on diverse mixtures and using regression to predict performance of unseen mixtures. \citet{wettig2025organize} demonstrate the value of fine-grained domain construction by applying RegMix to \textsc{WebOrganizer} topic and format labels, optimizing performance on MMLU \cite{hendrycks2021measuringmassivemultitasklanguage} and HellaSwag \cite{zellers2019hellaswagmachinereallyfinish}. While these efforts focus on topic, quality, and format as axes of variation, they do not consider narrative structure as a data dimension. Our work is complementary to this line of prior work by introducing \emph{narrative qualities} to the characterization of pretraining data.

\section{Our Annotation Framework}
\label{sec:schema}

The formalization of narrative communication as a multidimensional object of analysis dates back to the work of \citet{cc2b6d46-77fa-32b7-bc84-2ada91b51052} and the Russian formalists in the early twentieth century. Such work culminated in Gérard Genette's structuralist model of narrative discourse \cite{genette1983narrative}, later synthesized by \citet{herman2011basic}. 
The ``classic model'' of narratology was translated into computational frameworks by \citet{piper-etal-2021-narrative} and further refined in \citet{piper2023computational,hamilton2026narrabench}, resulting in \textsc{NarraBench}, a 50-dimensional benchmark for computational narrative understanding \cite{hamilton2026narrabench}.
Fundamental to all of these theoretical models is a tripartite framework that grounds narrative communication in three primary dimensions: agency, eventfulness, and world-building. 


The theoretical literature introduces a further distinction between a categorical and scalar understanding of narrative. Classical narratology largely operated on the understanding of narrative as a binary distinction. 
Later theory, dating back to the work of \citet{giora1994degrees}, has emphasized instead the concept of ``narrativity,'' that narrative communication exists along a spectrum of degree or intensity \cite{ochs2009living, pianzola2018looking, piper2022narrativity}.

By operationalizing these intersecting elements as interpretable rating dimensions for human annotators, we aim to bring narrative-theoretic concepts into contact with large-scale annotation. 
We provide an overview of our framework here, with full details of the annotation rating scales in App.~\ref{sec:feature_scales}.

\subsection{Agency}
\label{sec:agency}
 
Five dimensions capture how characters are represented as agents, rated on a 5-point Likert scale measuring centrality (1 = not central, 5 = extremely central) and drawing on Herman's account of experiential perspective in narrative \cite{doi:https://doi.org/10.1002/9781444305920.ch6}, which emphasizes that narratives convey ``what it's like'' to undergo events from a particular consciousness. Fig.~\ref{fig:agency_example} shows an example annotation.

\begin{figure}[t]
    \centering
    \small
    \setlength{\fboxsep}{8pt}
    \fbox{%
    \begin{minipage}{0.92\columnwidth}
    \parbox[t]{\linewidth}{\raggedright\itshape
    ``I walked into the interview room and
    \ulsolid{immediately noticed}{agencycolor} the panel was twice the size I expected.
    \uldash{My hands started shaking}{agencycolor}.
    I sat down, took a breath, and
    \uldot{somehow got through}{agencycolor} the first question.''
    } \\[6pt]
    
    \mbox{\legsolid{agencycolor}~Focalization} \quad
    \mbox{\legdash{agencycolor}~Emotion} \quad
    \mbox{\legdot{agencycolor}~Change of state } \\[8pt]
    \renewcommand{\arraystretch}{1.25}
    \begin{tabular}{@{} l c @{\hspace{8pt}} l c @{}}
    Focalization & \mbox{\score{4}{agencycolor}} & Change & \mbox{\score{3}{agencycolor}} \\
    Emotion & \mbox{\score{4}{agencycolor}} & Conflict & \mbox{\score{3}{agencycolor}} \\
    Cognition & \mbox{\score{3}{agencycolor}} & & \\
    \end{tabular}
    \end{minipage}%
    }
    \caption{``I [\ldots] immediately noticed'' signals focalization, ``my hands started shaking'' conveys emotion through observable behavior, and ``somehow got through'' implies a change of state from overwhelmed to managing. Scores reflect the passage as a whole, not individual phrases.}
    \label{fig:agency_example}
\end{figure}

\smallskip\noindent
\textbf{Focalization} captures the degree to which events are filtered through a specific character's perspective rather than described from an external vantage point, e.g., first-person narration with interiority, close third-person narration, or direct quotation of inner experience, and is inspired by \citet{doi:https://doi.org/10.1002/9781444305920.ch6}'s ``what it's like'' element of narratives.
 
\smallskip\noindent
\textbf{Internal emotion} captures the centrality of character's emotional states. Descriptions of emotional behavior 
can contribute to the score, but internal access raises it further. 
 
\smallskip\noindent
\textbf{Internal cognition} captures the centrality of a character's thoughts, reasoning, and motivations. This feature targets cognitive and reasoning interiority rather than perceptual or emotional interiority.
 
\smallskip\noindent
\textbf{Change of state} captures the centrality of a change in a character's condition, encompassing physical, psychological, relational, and existential transformations. 
In-progress, partially implied, or world-event-entailed changes (e.g., a company going bankrupt that necessarily affects a character) all count toward the score in varying degrees. 
 
\smallskip\noindent
\textbf{Conflict} captures the centrality of opposition or tension, encompassing interpersonal tension (character vs.\ character), internal struggle (character vs.\ self), opposition to institutions and physical environments (character vs.\ world). 
 
\subsection{Setting}
\label{sec:setting}

\begin{figure}[t]
    \centering
    \small
    \setlength{\fboxsep}{8pt}
    \fbox{%
    \begin{minipage}{0.92\columnwidth}
    \parbox[t]{\linewidth}{\raggedright\itshape
    ``I walked into
    \ulsolid{the interview room}{settingcolor} and
    immediately noticed
    \ulsolid{the panel}{settingcolor} was twice the size I expected.
    \uldash{My hands started shaking}{settingcolor}.
    I sat down, took a breath, and
    somehow got through the first question.''
    } \\[6pt]
    \mbox{\legsolid{settingcolor}~Concreteness} \quad
    \mbox{\legdash{settingcolor}~Sensory detail} \\[8pt]
    \renewcommand{\arraystretch}{1.25}
    \begin{tabular}{@{} l c @{\hspace{8pt}} l c @{}}
    Concreteness & \mbox{\score{3}{settingcolor}} & Spatial & \mbox{\score{2}{settingcolor}} \\
    Temporal & \mbox{\score{1}{settingcolor}} & Sensory & \mbox{\score{2}{settingcolor}} \\
    \end{tabular}
    \end{minipage}%
    }
    \caption{``The interview room'' and ``the panel'' provide concrete referents but are named rather than rendered. ``Hands started shaking'' is tactile but incidental. Temporal and spatial grounding score low: this could be taking place in any room at any time.}
    \label{fig:setting_example}
\end{figure}
 
This dimension measures how fully realized the storyworld is along four experiential dimensions, inspired by Herman's conception of narrative world-building \cite{doi:https://doi.org/10.1002/9781444305920.ch5}. Each dimension is rated on a 5-point Likert scale. Fig.~\ref{fig:setting_example} shows an example. 
 
\smallskip\noindent
\textbf{Concreteness} captures how tangible and non-abstract the language is. Concreteness is not the same as specificity: a passage can name specific entities and still be abstract if those entities are not rendered perceptually. This maps onto Herman's differentiation between \textit{existents} (things in the storyworld) and their \textit{rendering} \cite{doi:https://doi.org/10.1002/9781444305920.ch5}, and is grounded in psycholinguistic evidence linking concreteness to imageability and lexical processing \citep{richardson1975imagery}. 

\smallskip\noindent
\paragraph{Temporal grounding} captures how strongly the passage anchors the reader in a particular time, building on Gen\`{e}tte's analysis of narrative time \cite{genette1983narrative}. This manifests in two ways: \textit{historical grounding}, which locates the reader in a specific moment (a year, an era), and \textit{cyclical grounding}, which locates the reader within a recurring temporal structure (e.g., season, time of day).

\smallskip\noindent
\paragraph{Spatial grounding} captures how strongly the passage anchors the reader in a particular place. Spatial grounding manifests in two ways: \textit{geographic grounding} (a country, city, or named landmark) and \textit{proximate grounding} (a room, a building, a street). Geographic grounding is efficient and has the highest ceiling on its own; proximate grounding requires more rendering to score high. The two types work together to produce the highest scores. 

\smallskip\noindent
\paragraph{Sensory detail} captures how central sensory experience is to the text across modalities. This feature measures whether sensory experience is a prominent feature of the text. It is distinct from concreteness: a passage can be concrete without foregrounding any particular sense modality.

\subsection{Event Relations}
\label{sec:event_relations}

\begin{figure}[t]
    \centering
    \small
    \setlength{\fboxsep}{8pt}
    \fbox{%
    \begin{minipage}{0.92\columnwidth}
    \parbox[t]{\linewidth}{\raggedright\itshape
    ``I {\setulcolor{eventcolor}\ul{\textbf{walked} into the interview room}} and {\setulcolor{eventcolor}\ul{immediately \textbf{noticed}}} the panel was twice the size I expected. My hands started \textbf{shaking}. I \textbf{sat} down, took a \textbf{breath}, and somehow got through the first \textbf{question}.''
    } \\[6pt]
    {\color{eventcolor}\rule{10pt}{2pt}} Selected event spans (all event triggers in \textbf{bold}) \\[8pt]
    \renewcommand{\arraystretch}{1.25}
    \begin{tabular}{@{} l l @{}}
    \textit{\textbf{Temporal order}} & ``\textit{walked}" happens before ``\textit{noticed}" \\
    \textit{\textbf{Causal relation}} & ``\textit{walked}" enables ``\textit{noticed}" \\
    \end{tabular}
    \end{minipage}%
    }
    \caption{Two randomly selected adjacent event triggers are identified (``walked'' and ``noticed''). The event ``walked into the room'' precedes ``noticed the panel''. The events have a causal relationship: entering the room \textit{enables} the character to notice the panel. Event density counts all triggers in the passage, while temporal order and causal relation describe the selected pair.}
    \label{fig:event_example}
\end{figure}
 
Event relations capture the degree to which a text presents events in temporal sequence and links them through causal relationships. 
Event trigger detection, a requirement for event relation extraction, has a long history, from TimeBank \cite{pustejovsky2006timebank} to neural models trained on ACE \cite{doddington-etal-2004-automatic} and ERE \cite{wang2022mavenereunifiedlargescaledataset}. 

We define an event as a singular, bounded occurrence where something happens at a particular point in time, for which one can identify both \textit{what happened} and \textit{who or what it happened to}. Hypothetical, negated, and future events do not qualify. To measure event relations, we first identify event trigger spans using a \textsc{RoBERTa}-based event detection model fine-tuned on the event trigger data from \citet{sims-etal-2019-literary}. We validate the model's performance on our web-scale corpus before proceeding (achieves an F1 of 0.85, see App.~\ref{sec:event_span_detection}).
 
For \textbf{temporal ordering}, annotators indicate which event occurred first in time, selecting \textit{simultaneous} if the start points coincide, or selecting \textit{too hard to tell} if the text does not support a clear ordering. Following \citet{ning-etal-2018-multi}, we operationalize temporal order in terms of event start points rather than full event intervals. For \textbf{causal relation}, annotators choose among three categories: \textit{direct cause} (Event~1 is sufficient to produce Event~2), \textit{enablement} (Event~1 opens conditions for Event~2 without being sufficient), or \textit{not related} (no causal link is supported by the text or world knowledge). For downstream analysis, we collapse direct cause and enablement into a single \textit{causal} label, yielding a binary causal/not-causal distinction. We also calculate \textbf{event density}, the number of event triggers per token in the passage. Fig.~\ref{fig:event_example} shows an example of the event relation annotation with event triggers bolded.

\section{Sampling a Pretraining Corpus}
\label{sec:dataset}
 
\textsc{Dolma} \cite{dolma2024} is an open corpus of over 3~trillion tokens designed to support language model pretraining. We sample from twelve sources spanning web pages, news, encyclopedic text, books, and social media (Table~\ref{tab:sources}).
 
Raw \textsc{Dolma} documents present several challenges for narrative annotation. Full webpage text often includes navigation elements, lists of unrelated items, and fragmented or boilerplate content. Documents also vary enormously in length and register, from three-sentence forum posts to multi-thousand-word articles. Our sampling pipeline addresses these challenges in four steps. Additionally, the \textsc{Dolma} dataset is comprised of almost entirely English. We acknowledge this as a limitation, and conduct an initial distributional comparison between Arabic, German, French and English in App.~\ref{sec:multilingual_dist}.
We find strong correlations across these languages, but more work on multilingual narrative analysis is needed.
 
\paragraph{Step 1: Initial passage extraction.}
We draw $\sim$51M three-sentence passages from $\sim$15M unique documents across the raw \textsc{Dolma}~v1.7\footnote{\href{https://huggingface.co/datasets/allenai/dolma}{huggingface.co/datasets/allenai/dolma}} shards, allocating proportionally across sources according to the weights in Table~\ref{tab:sources}. Sentences are segmented using \textsc{NLTK} \texttt{sent\_tokenize}.
 
\paragraph{Step 2: Narrative scoring.}
We score each passage with a \textsc{RoBERTa}-based binary narrative classifier fine-tuned on data from \textsc{NarraDetect} \cite{piper-bagga-2025-narradetect} and \textsc{StorySeeker} \cite{antoniak-etal-2024-people}. The classifier returns a continuous confidence score $p \in [0,1]$. This score is used in Step~4 to focus the majority of final samples on passages with coherent narrative content.
 
\paragraph{Step 3: Topic classification.}
We assign topic and format labels to one chunk per document using the \textsc{WebOrganizer} topic and format classifier \cite{wettig2025organize}, which distinguishes 24 topics and formats. Topic labels are used in Step~4 to balance the final samples across subject domains, preventing high-frequency topics like news from completely dominating the final corpus. Format labels are used later for data analysis.
 
\paragraph{Step 4: Final samples.}
We draw two stratified samples from the scored and classified passage pool, both preserving the inter-source proportions in Table~\ref{tab:sources}. The first is the \textbf{gold dataset} which consists of 400 passages for human annotation. Of these, 85\% are drawn from passages with narrative scores $p > 0.50$ and 15\% are drawn without score filtering. For Common~Crawl sources, passages are further balanced across topics. We split these 400 passages into two non-overlapping gold sets for the agency and setting dimensions: one for validating LLM classifications (\textbf{gold split A}), and one held out for evaluating \textsc{NarraBert} classifications (\textbf{gold split B}). Because event-relation annotations require two confirmed event spans, event pairs are too sparse to split, so we evaluate event relations on the full annotated set at both validation stages.

The second is the full \textbf{\textsc{NarraDolma}} dataset which consists of $\sim$13M passages spanning $\sim$3.7M unique web documents. It follows the same stratified sampling procedure, with the unfiltered proportion shifted from 15\% to 25\% to increase coverage of non-narrative text. The raw \textsc{Dolma} source distribution is shown in Fig.~\ref{fig:narradolma_source_distribution} and the distribution of \textsc{NarraDolma} categories (with topics applied to Common Crawl sources) is shown in Fig.~\ref{fig:narradolma_combined_topic_source_dist}.

\section{Datasets}
\label{section-datasets}
 
\subsection{Human-Annotated Gold Dataset}
 
Annotations were collected using a custom \textsc{Potato}~\citep{pei-etal-2022-potato} annotation service (see App.\ref{sec:potato_ui}). One author annotated all three tasks for the full gold dataset ($N=400$). For verification, a second author and an additional annotator combined for $122$ overlapping labels for agency, $N=120$ for setting, and $N=275$ for events.

 

Human inter-annotator agreement is reasonable across all three tasks. For agency and setting, mean $\alpha = 0.75$ (range: $0.69$--$0.80$) and $0.69$ ($0.65$--$0.75$) respectively, with mean MAE of $0.61$ and $0.54$. For event relations, mean $\kappa = 0.68$ and mean F1 $= 0.90$. Per-dimension breakdowns are in App.~\ref{sec:agreement_scores}. 

\subsection{LLM-Labeled Dataset}
 
Following established validation-first workflows for LLM-assisted annotation \cite{pangakis2023automatedannotationgenerativeai}, we compare three models against human labels before selecting \textsc{Gemma} for large-scale labeling: \textsc{Claude Sonnet~4.6}, \textsc{Qwen3-235B-A22B}, and \textsc{Gemma4-31B}. 
 
\paragraph{LLM labeling at scale.} Agency and setting labels are produced by a single LLM call per passage. For event relations, we move beyond the single adjacent span pair used in validation to label relations for \emph{all} adjacent event pairs in a passage. We consider two passage-level scores: \textit{temporal sequencing} is the fraction of pairs with temporal relations; \textit{causal density} is the fraction of event pairs with causal relations. Both range from 0 to 1. Full prompts are in Fig.~\ref{fig:agency_prompt}, \ref{fig:setting_prompt}, and~\ref{fig:event_prompt}. The final LLM-labeled dataset contains 25K
passages, stratified by source and topic to preserve their original distribution.

\paragraph{LLM validation.} We validate all three models against gold split~A ($N=200$) for agency and setting, and against the full set of human event annotations for event relations. No single model dominates. For agency and setting, mean $\alpha$ across the three models is $0.71$ ($0.49$--$0.88$) with mean MAE of $0.53$ ($0.37$--$0.73$). For event relations, mean F1 is $0.78$ ($0.77$--$0.79$) and mean $\kappa$ is $0.56$ ($0.54$--$0.58$). We proceed with \textsc{Gemma} for large-scale labeling due to its cost effectiveness and sufficient performance. Full per-dimension and per-model agreement breakdowns are in App.~\ref{sec:agreement_scores}. 

\subsection{\texorpdfstring{\narradolma}{NarraDolma} and \texorpdfstring{\narrabert}{NarraBERT}}
\label{sec:narradolma_intro}

To scale narrative annotation beyond what LLM inference budgets allow, we use the \textsc{Gemma} labels to create \textsc{NarraBERT}, a pair of finetuned \textsc{RoBERTa} classifiers trained via knowledge distillation \cite{pangakis2024knowledgedistillationautomatedannotation}. This converts the LLM's per-passage judgments into lightweight models that can label millions of texts at low cost.

\paragraph{Training.} Both classifiers are finetuned on \textsc{Gemma} pseudo-labels, but on two separate 25K passage draws. For agency and setting, each training example is a passage with nine corresponding Likert-scale labels, drawn unweighted from the corpus. For event relations, each training example is a span-level event pair identified by the \citet{sims-etal-2019-literary} model, each pair with binary labels for temporal ordering and causal relation. Because only passages with at least two spans yield a pair (35\% of the passages) the second draw upsamples event-dense passages, giving 61K pairs rather than the 24K the unweighted draw produces at the same annotation cost. Drawing separately keeps the Likert labels on the corpus distribution. We frame the Likert dimensions as regression tasks and the event relation dimensions as binary classification tasks. \textsc{NarraBERT} consists of two \textsc{RoBERTa-base} encoders, one of which is a shared encoder with nine regression heads for the agency and setting dimensions and the other a dedicated encoder for event relations. Hyperparameters can be found in App.~\ref{sec:hyperparams}.

\paragraph{Validation against human annotations.} We validate \textsc{NarraBERT} against gold split~B ($N=200$) for agency and setting, and against the full set of human event annotations for event relations. For agency and setting, mean $\alpha = 0.68$ ($0.52$--$0.78$) with mean MAE of $0.57$ ($0.41$--$0.73$). For the two event relation features, mean F1 $= 0.75$ ($0.73$--$0.77$) vs.\ \textsc{Gemma}'s $0.78$ ($0.77$--$0.79$). Agreement is broadly comparable to the LLMs themselves. Full per-dimension breakdowns for each feature are in App.~\ref{sec:agreement_scores}.

\paragraph{Scaling up labels.}
We apply \textsc{NarraBERT} to the full \textsc{NarraDolma} corpus of $\sim$13M
passages across 12 Dolma sub-corpora. For agency and setting features, the shared encoder produces nine Likert scores per passage in a single forward pass. For event relations, we first run the \citet{sims-etal-2019-literary} event detector to identify event trigger spans in each passage, then apply the event relation encoder to classify every adjacent span pair which yield the temporal sequencing and causal density scores. The event detector itself also yields event density, the number of events per token. This produces a 12-dimensional narrative feature vector for each of the $\sim$13M passages, which together constitute the \textsc{NarraDolma} dataset. For all analyses in \S\ref{sec:data_analysis}, we aggregate these passage-level vectors to the document level by averaging across each document's passages, yielding one narrative vector per document ($\sim$3.7M documents), our \textbf{analysis corpus} of document-level \textbf{narrative profiles}. 

\section{The Narrative Landscape of Pretraining Data}
\label{sec:data_analysis}

\subsection{The Structure of Narrative Features}
\label{sec:narrative_dist}

Before comparing the narrative features across categories, we examine the distributions of the 12 narrative features across the full \textsc{NarraDolma} corpus (Fig.~\ref{fig:feature_distributions}). The five agency dimensions cluster near the floor ($\sim$1.0 on a 1--5 Likert scale) while the setting features are more spread out across the full range of scores. High agency features require text to have a perceiving consciousness which seems to be a property the \textsc{NarraDolma} corpus lacks, on average, compared to features related to events or setting. Temporal and spatial grounding are features most web text has to some degree since dates and place names are abundant.

\begin{figure*}[t]
    \centering
    \includegraphics[width=1\linewidth]{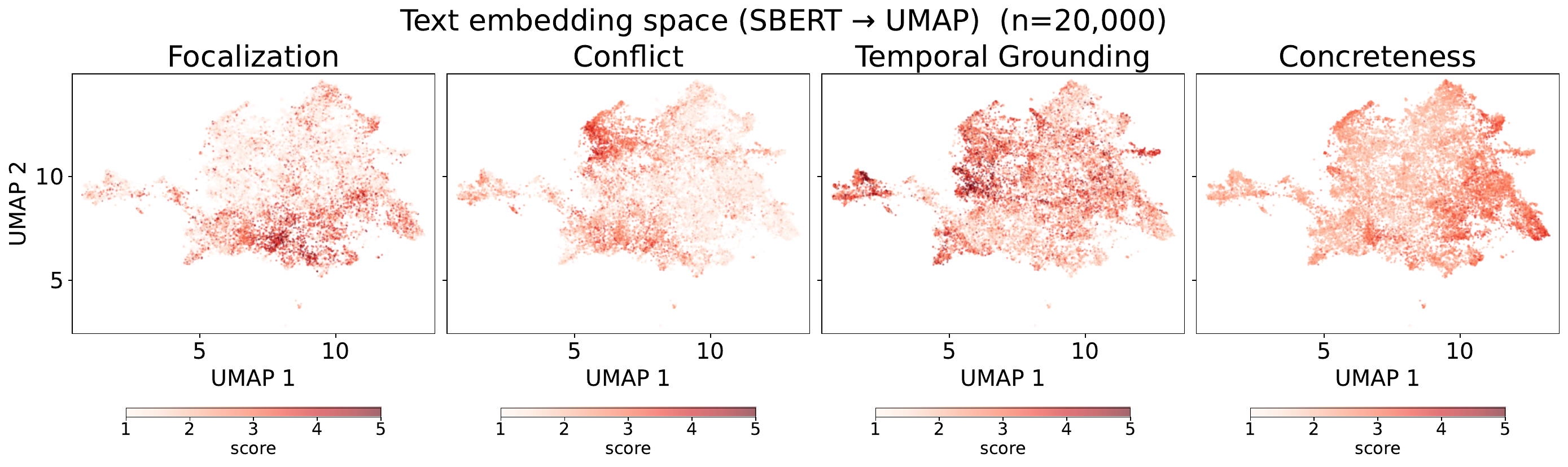}
    \caption{20K documents randomly sampled from \textsc{NarraDolma}, represented as SBERT embeddings and reduced to two UMAP dimensions, overlayed with focalization, conflict, temporal grounding, and concreteness scores. See the SBERT embedding space with annotated topic labels in Fig.~\ref{fig:umap_topic_centroids}.}
    \label{fig:umap_4_features}
\end{figure*}

To investigate whether the narrative content is recoverable from lexical content, we overlay feature scores onto a UMAP reduction of SBERT embeddings for 20K randomly sampled \textsc{NarraDolma} documents. Fig.~\ref{fig:umap_4_features} shows four of the narrative feature scores (focalization, conflict, temporal grounding, and concreteness). Each of the four plots show distinct patterns. If narrative features were a re-description of subject matter, we would expect documents nearby in embedding space to share narrative profiles. However, we see that each feature has a distinct pattern over identical points, with some narrative features more similar than others. Fig.~\ref{fig:umap_all_feature_overlay} shows all feature scores overlayed upon the reduced SBERT embeddings, and Fig.~\ref{fig:umap_topic_centroids} shows the WebOrganizer topic centroids labeled upon the SBERT embeddings.


\subsection{How Narrative Features Relate to Each Other}
\label{sec:feature_relations}

\begin{figure}[th]
    \centering
    \includegraphics[width=1\linewidth]{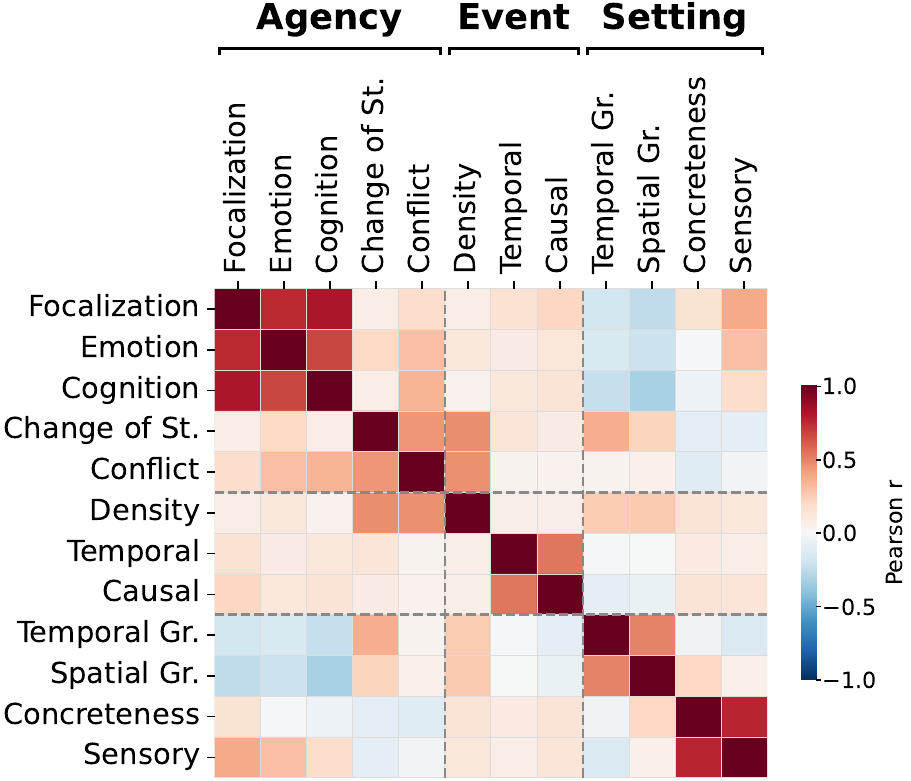}
    \caption{Pearson correlation of each feature pair, across our analysis corpus. Cells involving temporal sequencing or causal density use the 2.2M documents where those features are defined.}
    \label{fig:feature_correlation_heatmap_full}
\end{figure}

We expect some narrative features to be correlated, based both on our initial examinations of the data and on the theoretical overlap between subsets of the features.
We calculate the Pearson correlation between each of the 12 narrative features shown in Fig.~\ref{fig:feature_correlation_heatmap_full}. We see ``blocks" of high correlation where certain subsets of features are more highly correlated with each other than others. Focalization, emotion, and cognition form an \emph{interiority} block ($r = 0.75, p<0.005$, peaking at $0.82$ between focalization and cognition), and concreteness and sensory detail form a \emph{texture} block ($r = 0.76, p<0.005$). Three more blocks display moderate correlation: temporal and spatial grounding, temporal sequencing and causal density, and the looser grouping of change of state, conflict, and event density all cohere in the mid-$0.4$s to low-$0.5$s. Across blocks, interiority runs mildly against grounding ($r = -0.23, p<0.005$). 

To further explore this correlation we ran principal-component analysis (PCA) on the 10 non-event-relation features (App.~\ref{sec:pca}). The first three principal components (PCs) explain 71\% of the variance and correspond to interpretable axes that are similar to the correlation patterns. These dimensions can be used as summary axes for sampling, in cases where a coarser categorization is helpful.

\subsection{Where Narrativity Concentrates in Pretraining Data}
\label{sec:narr_concentration}

We next examine narrative profiles across \textsc{Dolma} sources, topics, and formats (\S\ref{sec:dataset}). The \textbf{\textsc{Dolma} source} is where the text comes from, while WebOrganizer~\citep{wettig2025organize} provides the \textbf{topic} (what the text is about) and the \textbf{format} (how the text is written). These categories have been used when curating pretraining data mixtures to improve performance on downstream tasks. But can sampling over these categories improve narrative coverage? How is narrative content distributed across these categories?

In the following results, temporal sequencing and causal density are averaged over the 2.2M documents where those features are defined, otherwise results are reported for all 3.7M documents in our analysis corpus.

\begin{figure}[th]
    \centering
    \includegraphics[width=1\linewidth]{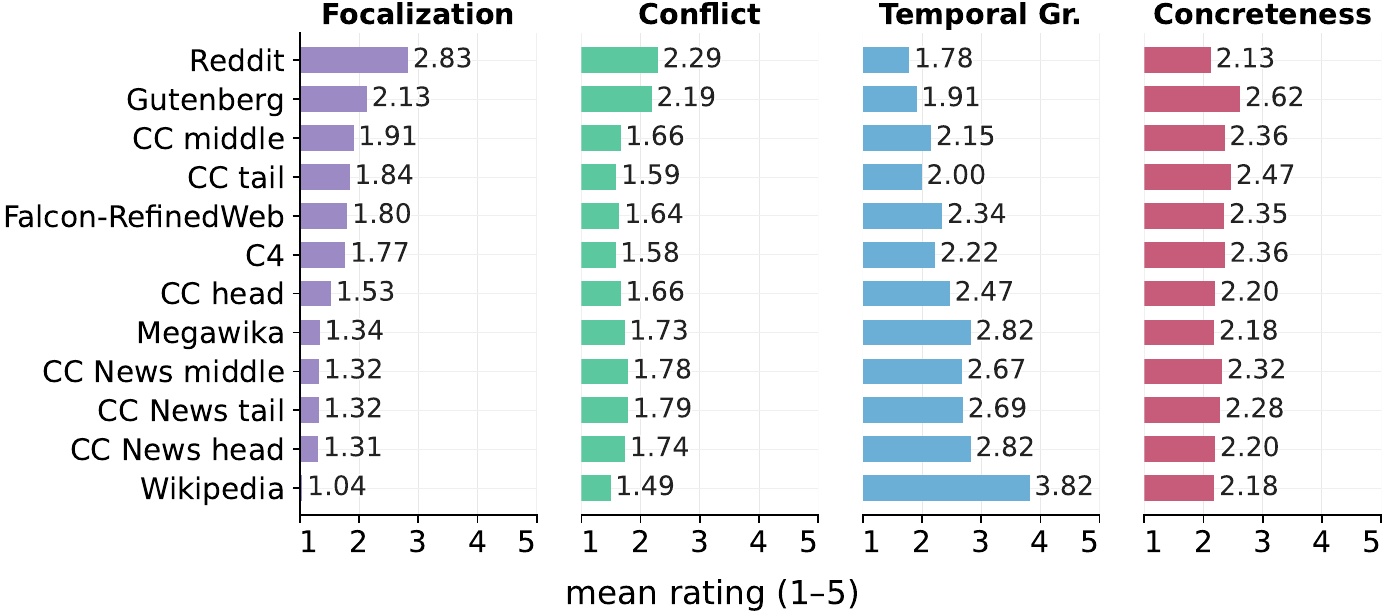}
    \caption{Bar plot of the focalization, conflict, temporal grounding and concreteness feature scores \textbf{by source}. All features are shown in App.~\ref{sec:bar_plots}.}
    \label{fig:source_bar_plot_interiority}
\end{figure}

\paragraph{\textsc{Dolma} Sources.} Fig.~\ref{fig:source_bar_plot_interiority} shows the mean scores for focalization, conflict, temporal grounding and concreteness by source. Three of the more specifically-curated sources (Reddit, Gutenberg, and Wikipedia) show distinct narrative profiles. Reddit and Gutenberg score high on focalization and conflict, and lower on temporal grounding. Wikipedia shows the exact opposite narrative profile. Fig.~\ref{fig:subcorpus_mean_heatmap} shows the full mean $z$-scored narrative features by \textsc{Dolma} source.

\begin{figure}[th]
    \centering
    \includegraphics[width=1\linewidth]{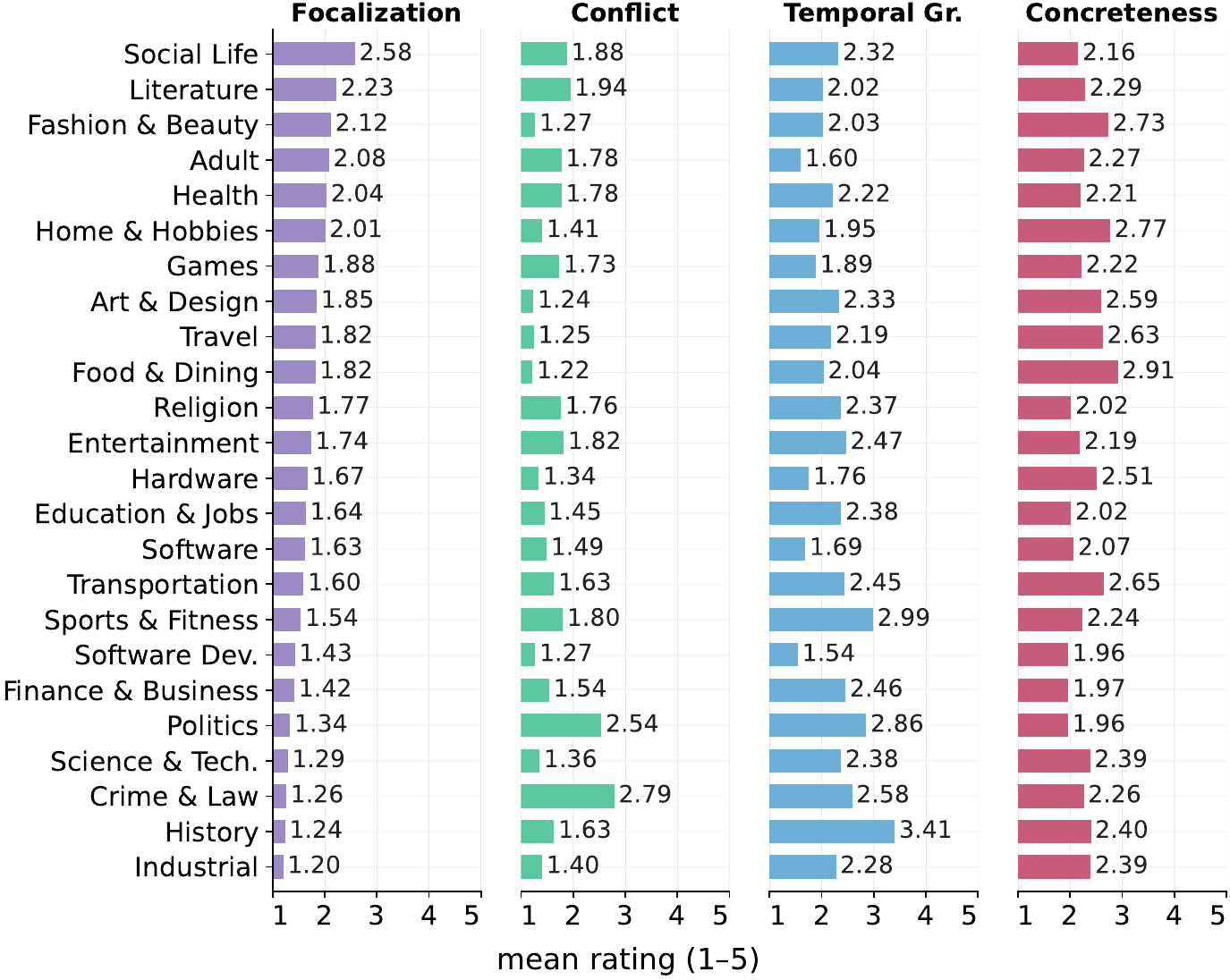}
    \caption{Bar plot of the focalization, conflict, temporal grounding and concreteness feature scores \textbf{by topic}. All features are shown in App.~\ref{sec:bar_plots}.}
    \label{fig:topic_bar_plot_interiority}
\end{figure}

\paragraph{Topics.} Fig.~\ref{fig:topic_bar_plot_interiority} shows the mean scores for focalization, conflict, temporal grounding and concreteness by topic. The history topic scores high on temporal grounding but low on focalization, while social life and literature are the exact opposite. Crime \& law scores high on conflict but quite low on focalization. Fig.~\ref{fig:topic_mean_heatmap} shows the mean $z$-scored for the full feature set by topic. 

\begin{figure}[t]
    \centering
    \includegraphics[width=1\linewidth]{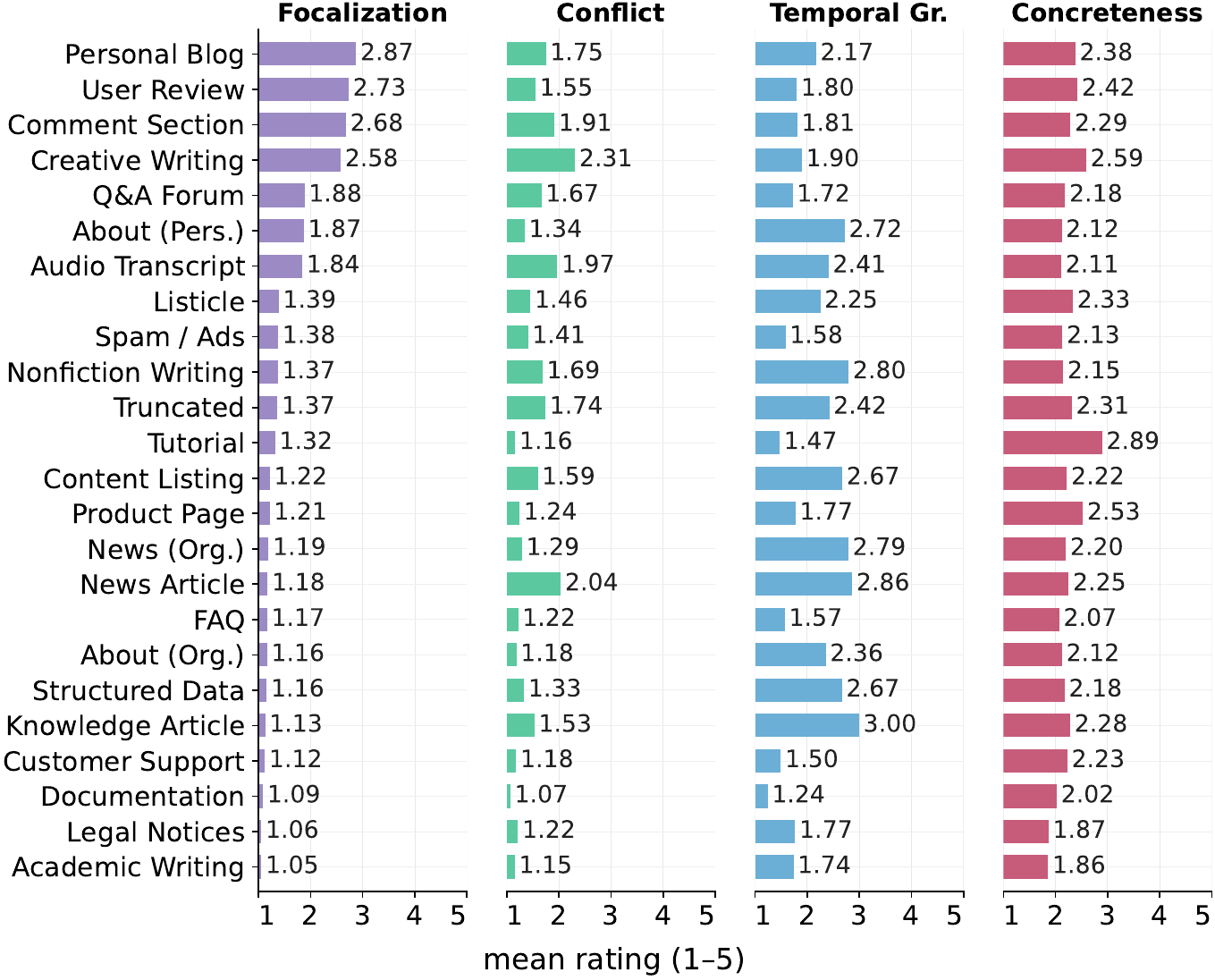}
    \caption{Bar plot of the focalization, conflict, temporal grounding and concreteness feature scores \textbf{by format}. All features are shown in App.~\ref{sec:bar_plots}.}
    \label{fig:format_bar_plot_interiority}
\end{figure}

\paragraph{Formats.} Fig.~\ref{fig:format_bar_plot_interiority} shows the mean scores for focalization, conflict, temporal grounding and concreteness by format. Narrative is a property of how a text is written (as opposed to what a text is about), and the format labels are the closest of the three categories (source, topic, and format) to describing this. Creative writing, personal blog, user review, and comment section score high on focalization. Tutorial scores high for concreteness but low for most other features. Fig.~\ref{fig:format_mean_heatmap} shows the full mean $z$-scored features by format.

\paragraph{Variation within categories.} The mean feature scores describe where the categories sit on average, but categories may also display high or lower variation. The within-category standard deviation (SD) in corpus SD units is 0.96 for \textsc{Dolma} source, 0.92 for topic, and 0.89 for format. A SD of 1 means a category is exactly as internally varied as the whole \textsc{NarraDolma} corpus. The more factual and technical subcategories like the Wikipedia \textbf{source} (0.68) and the documentation \textbf{format} (0.31) have comparably lower SDs. The more creative subcategories like the Reddit \textbf{source} (1.09) and the creative writing \textbf{format} (1.01) have higher SDs. The full set of SD values across all categories and features is in App.~\ref{sec:feature_var}. This is meaningful because the categories a model practitioner would upweight for more narrative text are precisely the ones with the widest spread of narrative features. Filtering \textit{within} these categories is necessary in order to effectively upweight specific narrative profiles.

\section{Discussion}
\label{sec:discussion}

Our analysis reveals three core empirical findings. 
First, we find that narrative can be measured as a multidimensional, continuous spectrum rather than a binary property.
Second, the narrative features are unequally distributed across categories in ways that current curation practices neither measure nor account for. 
Third, narrative variation within categories is substantial, 
indicating that source-level, topic-level, and format-level labels are too coarse to capture narrative diversity. Upweighting ``narrative'' sources would not uniformly increase all narrative qualities.
Together these results position narrative structure as a measurable, multidimensional property of web text rather than a single ``narrativity'' quantity. 

These findings have direct implications for data curation. Source-level mixing decisions, as currently practiced, cannot account for the multidimensional structure of narrative in pretraining data. A training mixture heavy on Reddit would amplify interiority while starving the model of storyworld texture and event structure; a mixture heavy on Wikipedia would do the reverse. Narrative-aware curation requires understanding which axes a given source contributes to, not just whether it is ``narrative'' in aggregate.

Looking forward, this work opens several directions. Controlled data mixing experiments using frameworks such as RegMix \cite{liu2025regmixdatamixtureregression} could establish direct causal links between pretraining narrative composition and downstream narrative capability. Analysis of intermediate training checkpoints could reveal when and how narrative competencies emerge during pretraining. 
Such audits could also support safety research about the narratives influencing model outputs, whether those narratives are desirable, and how they affect the stories that many people, including children, are already generating via popular chatbots.

Our framework, datasets, and models provide a foundation for these investigations and for treating narrative structure as a primary dimension of pretraining data composition, complementing existing axes of quality, toxicity, and topic distribution.

\section{Conclusion}
\label{sec:conclusion}

In this work we presented \textsc{NarraDolma}, a large-scale corpus characterizing narrative structure in pretraining data, together with \textsc{NarraBERT}, a pair of efficient classifiers for its 11 narrative dimensions. We developed an annotation framework grounded in narrative theory, validated through human annotation and LLM-assisted labeling, and applied it to produce narrative feature vectors for $\sim$13M passages across $\sim$3.7M documents.

\section{Limitations}

Our study has several limitations. First, \textsc{NarraDolma} is a stratified subsample ($\sim$13M passages) of a $>$3-trillion-token corpus, and it deliberately over-represents narrative content. The relative structural patterns we report are therefore informative but absolute prevalence figures do not transfer to raw \textsc{Dolma}. 

Second, human annotation was conducted on 400 passages, a small sample relative to the diversity of web text. This constraint reflects the fine-grained nature of our framework. Each passage requires judgment across 11 dimensions, and the iterative calibration process spanned several months. Primary annotation that was used for model validation was conducted by one author. Some agreement scores between annotators remain modest suggesting some of these dimensions are intrinsically difficult regardless of annotator expertise.

Third, the gold sets for agency and setting validation were non-overlapping, but event annotations were too sparse to split, so event evaluation is not fully independent across the LLM and classifier validation stages.

Fourth, our analysis is limited to English text and to a single theoretical operationalization of narrative. Other languages, narrative traditions, and frameworks may surface different structure. We do conduct an initial distribution comparison between English, Arabic, German, and French detailed in App.~\ref{sec:multilingual_dist}, but leave more robust analysis for future work.

Finally, we do not establish direct causal links between pretraining narrative composition and downstream model behavior. Our focus is on characterizing and mapping narrative structure in pretraining data, which is challenging and required designing new sampling, annotation, and analysis pipelines. Experiments such as controlled data remixing or checkpoint analysis remain important directions for future work.

\section*{Ethical Considerations}

This study relies on the open English language pretraining dataset \textsc{Dolma}, released with the Open Data Commons Attribution License (ODC-By).\footnote{\href{https://opendatacommons.org/licenses/by/1-0/}{https://opendatacommons.org/licenses/by/1-0/}}
This dataset includes massive amounts of web scraped data, including toxic, explicit, and personal data posted publicly to the internet.
We avoid highlighting such passages in this paper, but we did not avoid them during annotation, as the narrative qualities of this data may have significant implications for safe story generation.
We release the human-, LLM-, and \textsc{NarraBERT}-labeled datasets, which include the sampled portion of text along with the Dolma unique ID to allow for future rehydrating.
As the data authors could not have an expectation of privacy, we determined that review for this study was not required by our institution's Internal Review Board (IRB). 

\paragraph{Potential Risks}
Because \textsc{NarraDolma} includes both passage text and narrative feature annotations, it may make it easier to identify, filter, or prioritize web passages with particular narrative qualities, including emotionally intense, personal, violent, or explicit narratives. This could be misused to construct training mixtures that overrepresent sensational, manipulative, or harmful narrative styles, or to retrieve sensitive personal disclosures from publicly available web text. These risks are heightened because \textsc{Dolma} contains web-scraped data, including toxic, explicit, and personal content. 
However, releasing our annotations supports future research into the factors influencing story generation, which is a very popular task for users of modern chatbots and therefore of interest for safety audits.
Given the growing prominence of these tools and their effects on society, we believe studies like ours can contribute to better understandings, for researchers and the public. 

\paragraph{Use of AI Assistance}
We used AI assistants for limited support during coding, analysis, and writing. All data decisions, annotations, citations, statistical analyses, interpretations, and final writing decisions were reviewed and made by the authors.

\section*{Acknowledgments}

We thank our annotators for their careful work annotating the gold dataset, often across difficult and ambiguous passages, and for the many discussions that helped refine the annotation guidelines. We are grateful to our colleagues for feedback on earlier drafts of this work, and to members of the NLP Group at the University of Colorado Boulder for valuable discussion throughout the project. We also thank the anonymous reviewers for their constructive comments. Lastly we thank Rohan Das, Matt Pauk, Advait Deshmukh, and Uma Gunturi for their help with developing the annotation framework.

Computational resources were provided by the University of Colorado Boulder Research Computing group, including the Blanca condo cluster.

\newpage

\bibliography{anthology,custom}
\bibliographystyle{acl_natbib}

\appendix

\renewcommand{\thefigure}{A\arabic{figure}}
\renewcommand{\thetable}{A\arabic{table}}
\setcounter{figure}{0}
\setcounter{table}{0}
\clearpage
\section{Datasets \& Validation}

This appendix reports the validation of \textsc{NarraBert}: gold-set label distributions, inter-annotator agreement, and model agreement at each validation stage.

\begin{figure*}[th]
    \centering
    \includegraphics[width=1\linewidth]{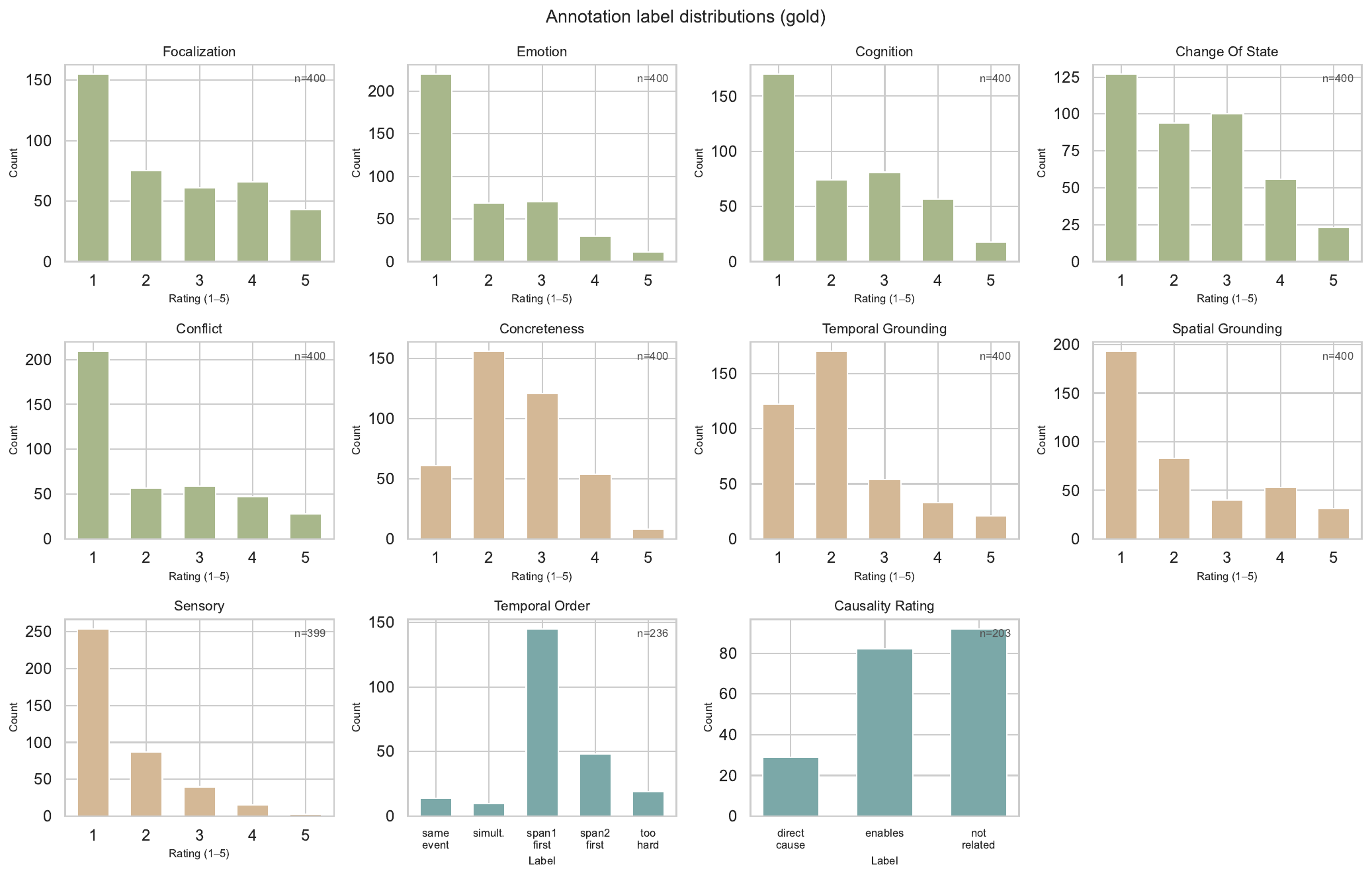}
    \caption{Distribution of gold annotated labels across the 11 manually-annotated features.}
    \label{fig:combined_gold_label_distributions}
\end{figure*}

\begin{figure*}[t]
    \centering
    \includegraphics[width=1\linewidth]{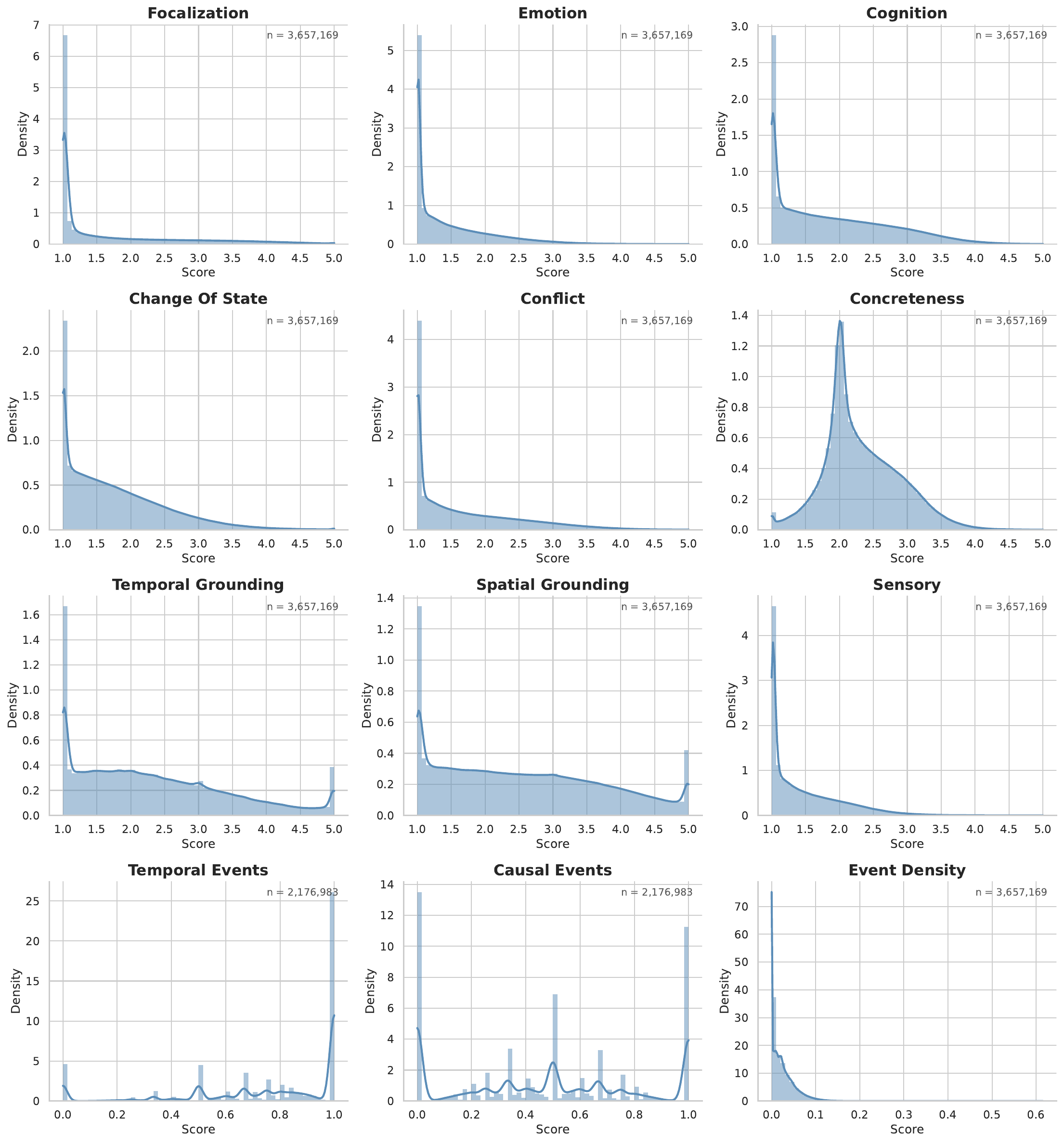}
    \caption{Distributions of each narrative feature in the full \narradolma corpus. The $n$ in the upper right corner of each plot indicates the number of documents shown in the distribution.}
    \label{fig:feature_distributions}
\end{figure*}

Figure~\ref{fig:combined_gold_label_distributions} shows the distributions of gold annotated labels across the 11 manually-annotated features. Figure~\ref{fig:feature_distributions} shows the distributions of the full \textsc{NarraDolma} corpus labels.

\clearpage
\subsection{Agreement Scores}
\label{sec:agreement_scores}

Tables~\ref{tab:app_human_iaa}, \ref{tab:app_llm_validation}, and \ref{tab:app_bert_validation} report full per-dimension agreement scores for each validation stage. Table~\ref{tab:app_human_iaa} reports human inter-annotator agreement on the 120 and 122-passage overlap set for agency and setting and on the full event annotation set. Agency and setting dimensions report MAE and Krippendorff's $\alpha$, while event relations report Cohen's $\kappa$ and macro F1. Table~\ref{tab:app_llm_validation} reports agreement between each of the three LLMs and gold split A ($N=200$ for agency and setting, full event set for event relations), with the best-performing model per dimension bolded. Table~\ref{tab:app_bert_validation} reports \textsc{NarraBert} agreement against the held-out gold split B ($N=200$ for agency and setting, full event set for event relations), using annotations not seen during LLM validation.

\begin{table}[th]
    \centering
    \small
    \begin{tabular}{llcccc}
    \toprule
    \textbf{Dimension} & \textbf{N} & \textbf{MAE} & \textbf{$\alpha$} & \textbf{$\kappa$} & \textbf{F1} \\
    \midrule
    \multicolumn{6}{l}{\textit{Agency}} \\
    \quad Focalization    & 122 & 0.639 & 0.795 & --- & --- \\
    \quad Emotion         & 122 & 0.541 & 0.709 & --- & --- \\
    \quad Cognition       & 122 & 0.615 & 0.776 & --- & --- \\
    \quad Change of state & 122 & 0.697 & 0.691 & --- & --- \\
    \quad Conflict        & 122 & 0.549 & 0.768 & --- & --- \\
    \midrule
    \multicolumn{6}{l}{\textit{Setting}} \\
    \quad Concreteness    & 120 & 0.647 & 0.666 & --- & --- \\
    \quad Temporal gr.    & 120 & 0.550 & 0.713 & --- & --- \\
    \quad Spatial gr.     & 120 & 0.583 & 0.747 & --- & --- \\
    \quad Sensory         & 120 & 0.392 & 0.649 & --- & --- \\
    \midrule
    \multicolumn{6}{l}{\textit{Event}} \\
    \quad Is span event        & 550 & --- & --- & 0.798 & 0.921 \\
    \quad Temporal        & 123 & --- & --- & 0.645 & 0.891 \\
    \quad Causality       & 82  & --- & --- & 0.608 & 0.901 \\
    \bottomrule
    \end{tabular}
    \caption{Human inter-annotator agreement. Agency and setting report MAE and Krippendorff's $\alpha$; event relations report Cohen's $\kappa$ and F1. For event annotation, annotators are asked to determine whether pre-highlighted event spans (detailed in App. \ref{sec:event_span_detection}) are events. If both pre-highlighted spans are events, annotators label temporal and causal relations.}
    \label{tab:app_human_iaa}
\end{table}

\begin{table*}[th]
    \centering
    \small
    \begin{tabular}{llccccccccc}
    \toprule
    & & \multicolumn{3}{c}{\textbf{Sonnet}} & \multicolumn{3}{c}{\textbf{Qwen3}} & \multicolumn{3}{c}{\textbf{Gemma4}} \\
    \cmidrule(lr){3-5} \cmidrule(lr){6-8} \cmidrule(lr){9-11}
    \textbf{Dimension} & \textbf{N} & \textbf{MAE} & \textbf{$\alpha$/$\kappa$} & \textbf{F1} & \textbf{MAE} & \textbf{$\alpha$/$\kappa$} & \textbf{F1} & \textbf{MAE} & \textbf{$\alpha$/$\kappa$} & \textbf{F1} \\
    \midrule
    \multicolumn{11}{l}{\textit{Agency ($\alpha$)}} \\
    \quad Focalization    & 200 & 0.605 & 0.760 & --- & 0.616 & 0.762 & --- & \textbf{0.597} & \textbf{0.764} & --- \\
    \quad Emotion         & 200 & 0.505 & 0.713 & --- & 0.453 & 0.694 & --- & \textbf{0.436} & \textbf{0.749} & --- \\
    \quad Cognition       & 200 & \textbf{0.545} & 0.737 & --- & 0.646 & 0.678 & --- & 0.559 & \textbf{0.753} & --- \\
    \quad Change of st.   & 200 & \textbf{0.565} & \textbf{0.739} & --- & 0.726 & 0.634 & --- & 0.657 & 0.703 & --- \\
    \quad Conflict        & 200 & \textbf{0.580} & \textbf{0.753} & --- & 0.637 & 0.644 & --- & 0.594 & 0.748 & --- \\
    \midrule
    \multicolumn{11}{l}{\textit{Setting ($\alpha$)}} \\
    \quad Concreteness    & 200 & \textbf{0.445} & \textbf{0.651} & --- & 0.610 & 0.602 & --- & 0.516 & 0.628 & --- \\
    \quad Temporal gr.    & 200 & \textbf{0.425} & \textbf{0.765} & --- & 0.506 & 0.720 & --- & 0.536 & 0.728 & --- \\
    \quad Spatial gr.     & 200 & 0.445 & 0.813 & --- & \textbf{0.378} & \textbf{0.883} & --- & 0.651 & 0.744 & --- \\
    \quad Sensory         & 200 & \textbf{0.370} & 0.574 & --- & 0.425 & 0.486 & --- & 0.384 & \textbf{0.639} & --- \\
    \midrule
    \multicolumn{11}{l}{\textit{Event ($\kappa$)}} \\
    \quad Temporal        & 218 & --- & \textbf{0.556} & \textbf{0.778} & --- & --- & --- & --- & 0.539 & 0.769 \\
    \quad Causality       & 180 & --- & 0.550 & 0.775 & --- & --- & --- & --- & \textbf{0.575} & \textbf{0.786} \\
    \bottomrule
    \end{tabular}
    \caption{LLM validation against split A of the gold annotations. Agency and setting report MAE and $\alpha$; event relations report $\kappa$ and F1. Qwen3 was not evaluated on event relations. Best per dimension in bold (lower is better for MAE, higher for all others).}
    \label{tab:app_llm_validation}
\end{table*}

\begin{table}[th]
    \centering
    \small
    \begin{tabular}{llccc}
    \toprule
    \textbf{Dimension} & \textbf{N} & \textbf{MAE} & \textbf{$\alpha$} & \textbf{F1} \\
    \midrule
    \multicolumn{5}{l}{\textit{Agency}} \\
    \quad Focalization    & 200 & 0.559 & 0.782 & --- \\
    \quad Emotion         & 200 & 0.432 & 0.712 & --- \\
    \quad Cognition       & 200 & 0.615 & 0.721 & --- \\
    \quad Change of st.   & 200 & 0.734 & 0.616 & --- \\
    \quad Conflict        & 200 & 0.538 & 0.719 & --- \\
    \midrule
    \multicolumn{5}{l}{\textit{Setting}} \\
    \quad Concreteness    & 200 & 0.586 & 0.656 & --- \\
    \quad Temporal gr.    & 200 & 0.570 & 0.727 & --- \\
    \quad Spatial gr.     & 200 & 0.690 & 0.659 & --- \\
    \quad Sensory         & 200 & 0.405 & 0.515 & --- \\
    \midrule
    \multicolumn{5}{l}{\textit{Event}} \\
    \quad Temporal        & 236 & --- & --- & 0.766 \\
    \quad Causality       & 193 & --- & --- & 0.727 \\
    \bottomrule
    \end{tabular}
    \caption{\narrabert validation against split B of the gold annotations. Agency and setting report MAE and $\alpha$; event relations report F1. Validated against held-out human annotations not used during LLM validation.}
    \label{tab:app_bert_validation}
\end{table}
\clearpage
\section{Analysis Details \& Additional Plots}
\label{sec:add_analysis}

This appendix expands the corpus analysis.

\subsection{Bar Plot Visuals}
\label{sec:bar_plots}

\begin{figure*}[t]
    \centering
    \includegraphics[width=1\linewidth]{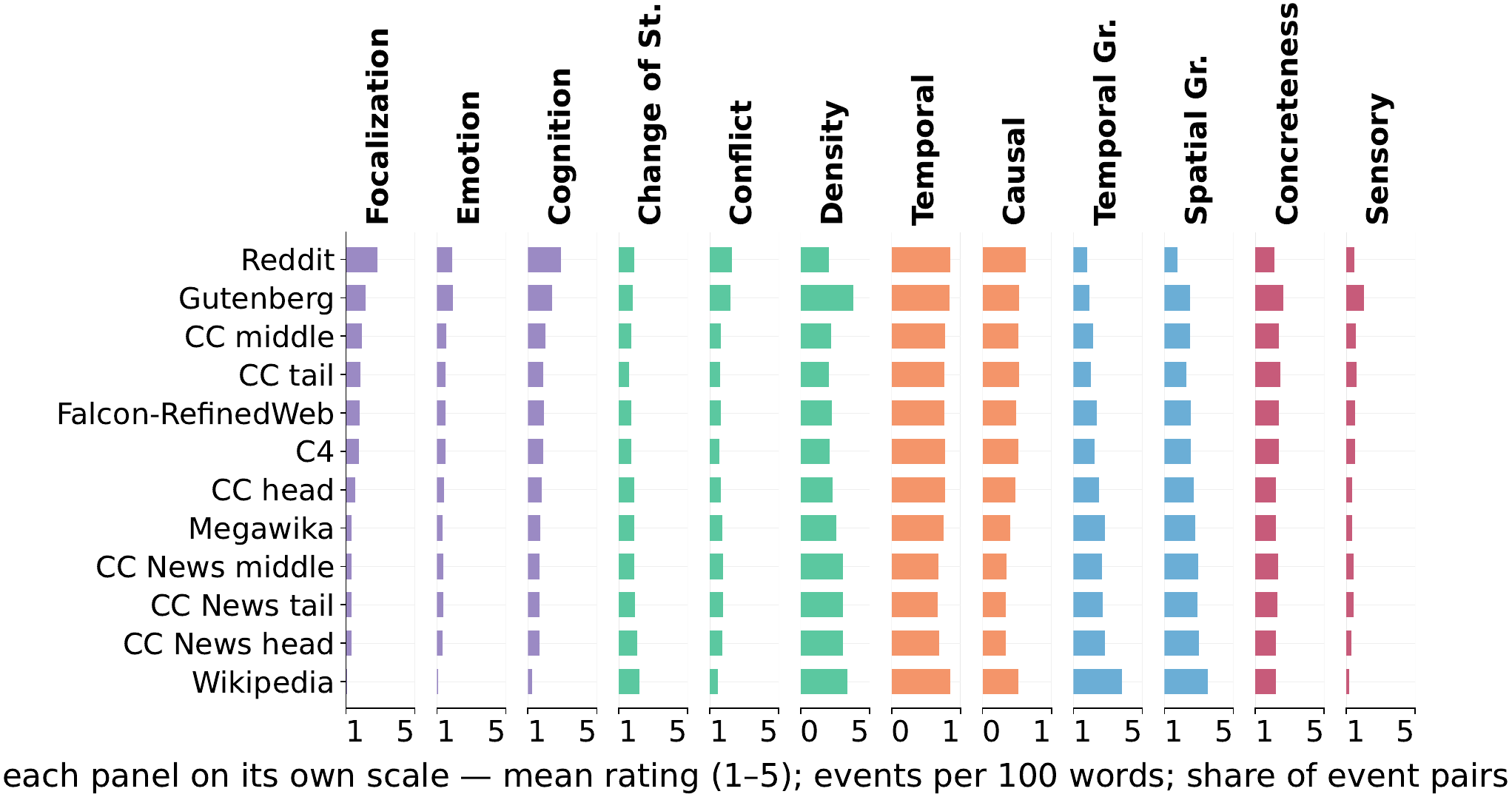}
    \caption{The mean of each feature across the full \textsc{NarraDolma} corpus by \textbf{source}. Colored by the five covarying ``blocks" described in Section~\ref{sec:feature_relations}.}
    \label{fig:source_bar_plot}
\end{figure*}

\begin{figure*}[t]
    \centering
    \includegraphics[width=1\linewidth]{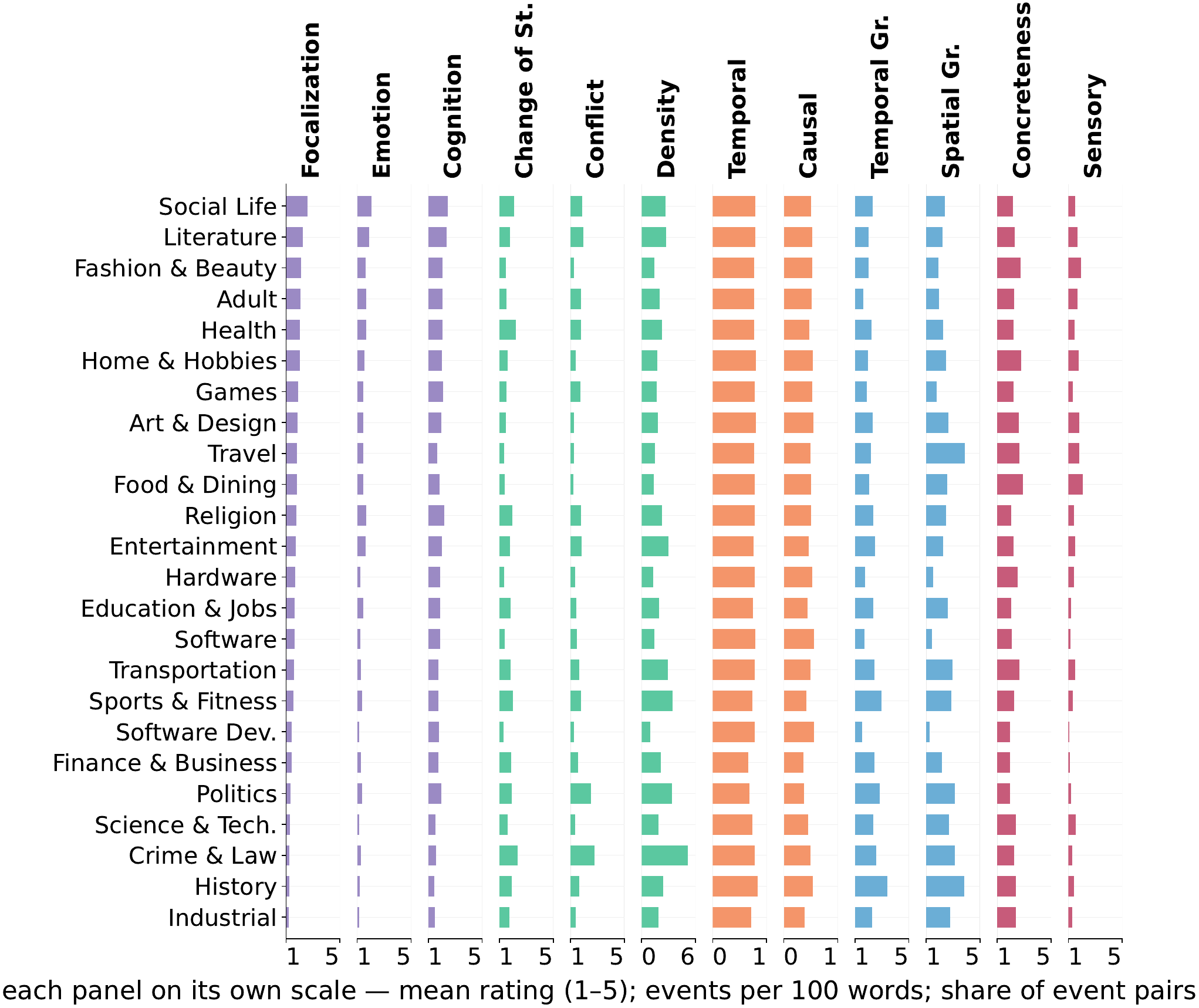}
    \caption{The mean of each feature across the full \textsc{NarraDolma} corpus by \textbf{topic}. Colored by the five covarying ``blocks" described in Section~\ref{sec:feature_relations}.}
    \label{fig:topic_bar_plot}
\end{figure*}

\begin{figure*}[t]
    \centering
    \includegraphics[width=1\linewidth]{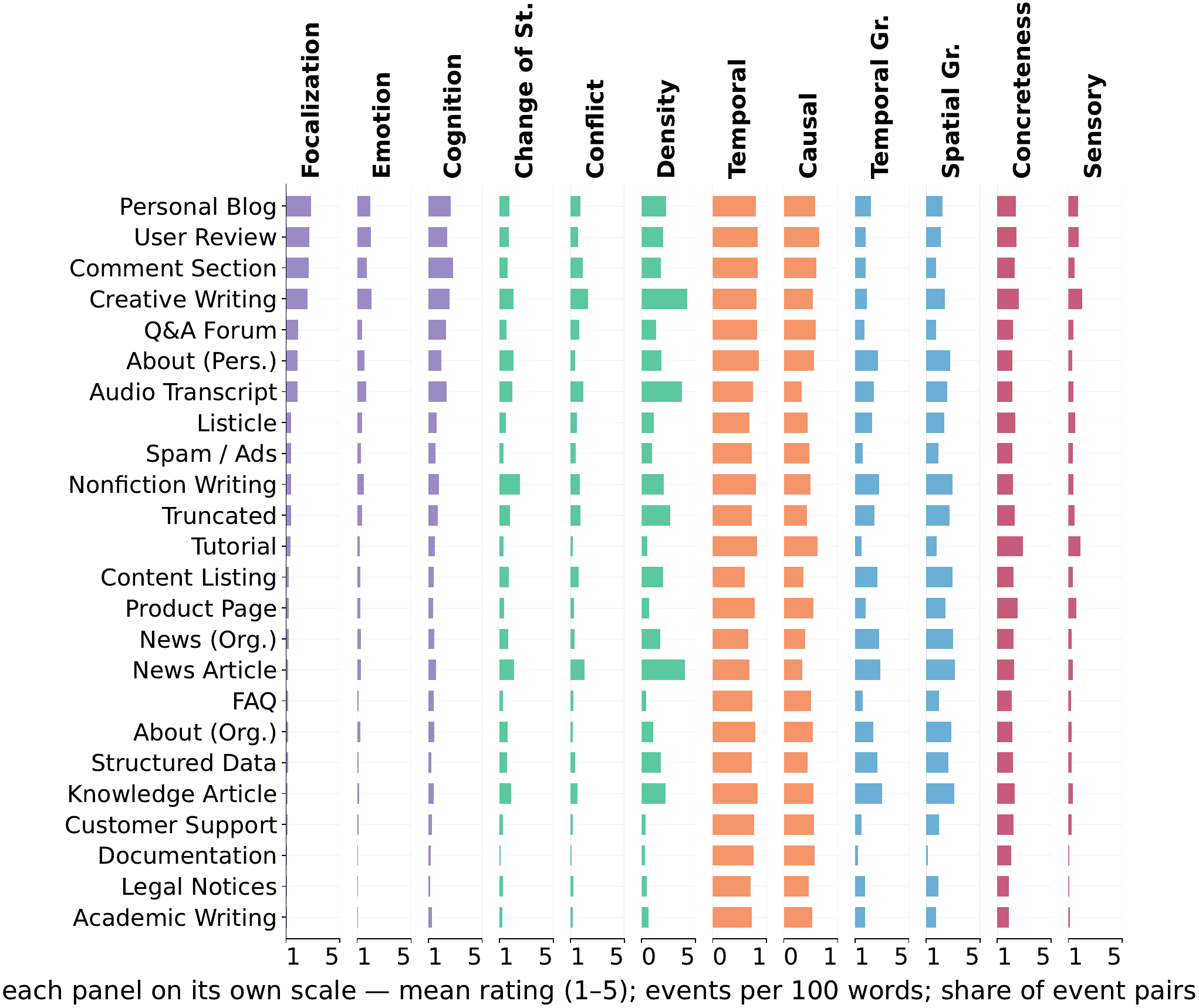}
    \caption{The mean of each feature across the full \textsc{NarraDolma} corpus by \textbf{format}. Colored by the five covarying ``blocks" described in Section~\ref{sec:feature_relations}.}
    \label{fig:format_bar_plot}
\end{figure*}

Figures~\ref{fig:source_bar_plot}, \ref{fig:topic_bar_plot}, and \ref{fig:format_bar_plot} show bar plots of the feature scores by \textsc{Dolma} source, topic, and format respectively.

\subsection{UMAP Visuals}
\label{sec:add_umap}

\begin{figure*}[t]
    \centering
    \includegraphics[width=1\linewidth]{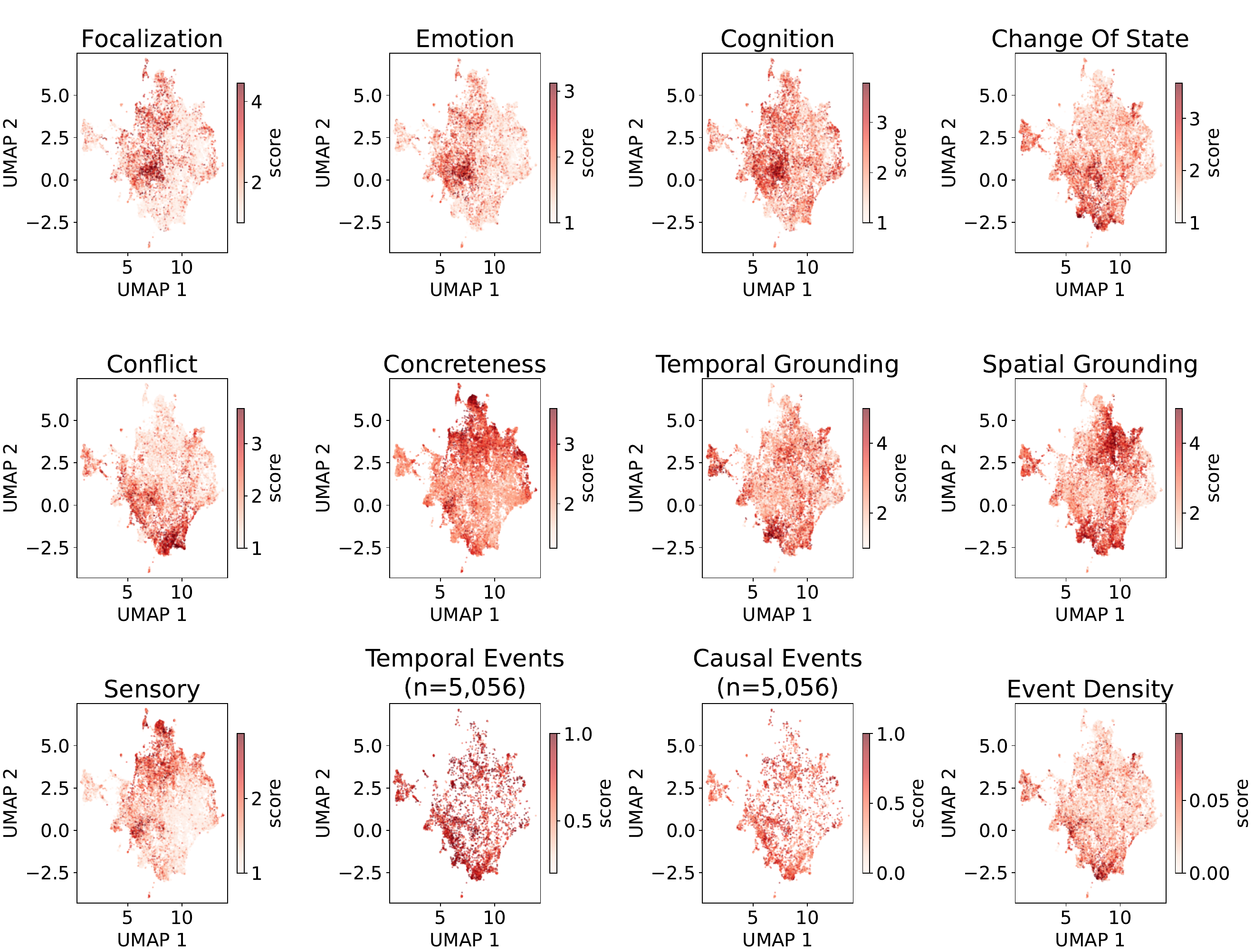}
    \caption{UMAP reduction of SBERT embeddings for 20,000 randomly sampled documents, colored by narrative feature scores.}
    \label{fig:umap_all_feature_overlay}
\end{figure*}

\begin{figure*}[t]
    \centering
    \includegraphics[width=1\linewidth]{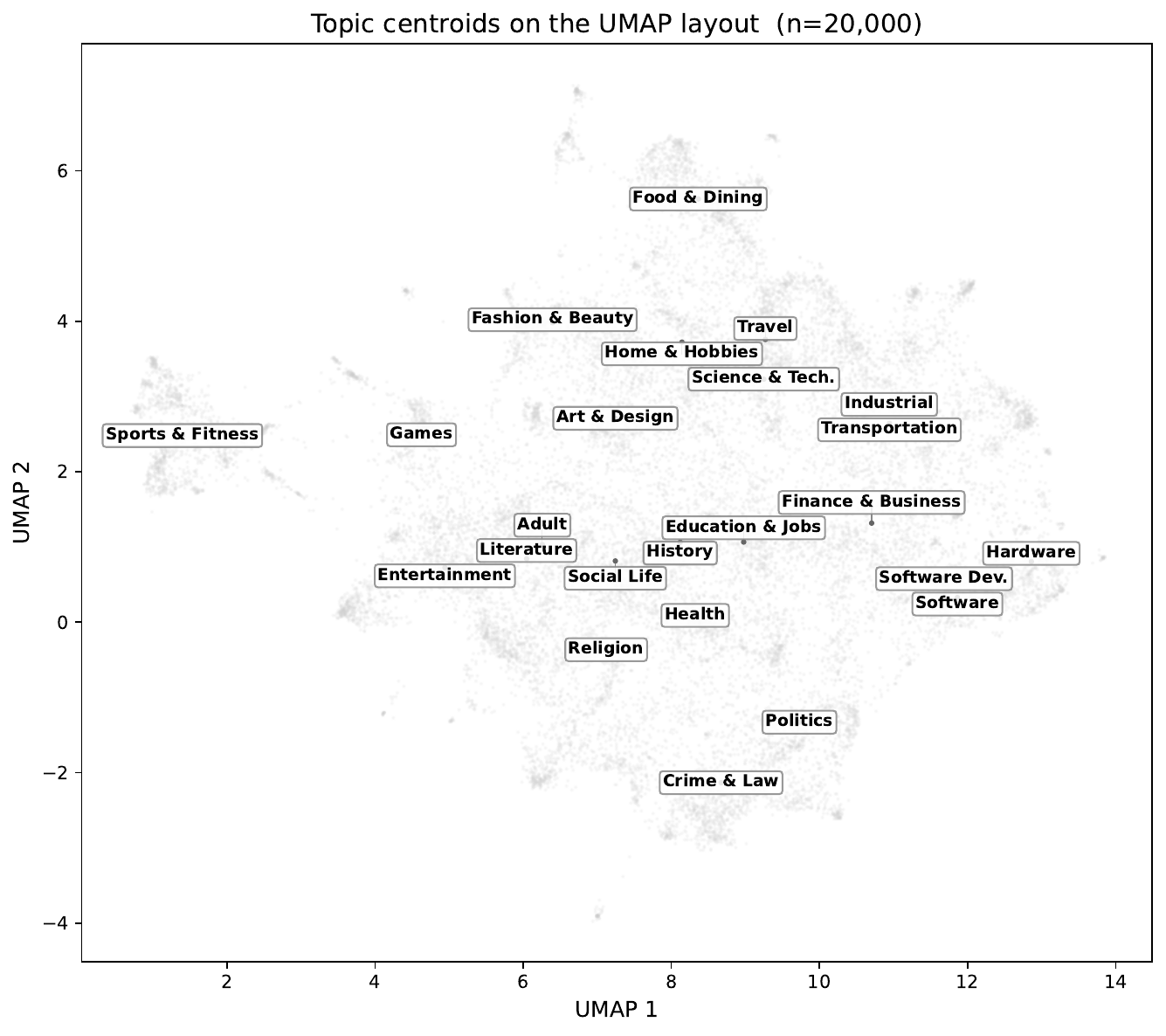}
    \caption{UMAP reduction of SBERT embeddings for 20,000 randomly sampled documents, annotated with the topic centroids from WebOrganizer labels.}
    \label{fig:umap_topic_centroids}
\end{figure*}

Figure~\ref{fig:umap_all_feature_overlay} shows all 12 feature scores overlayed upon a UMAP reduction of 20,000 randomly sampled passages. Figure~\ref{fig:umap_topic_centroids} shows the topic centroids overlayed upon the SBERT embeddings.

\clearpage

\subsection{PCA Details}
\label{sec:pca}
\begin{figure}[th]
    \centering
    \includegraphics[width=1\linewidth]{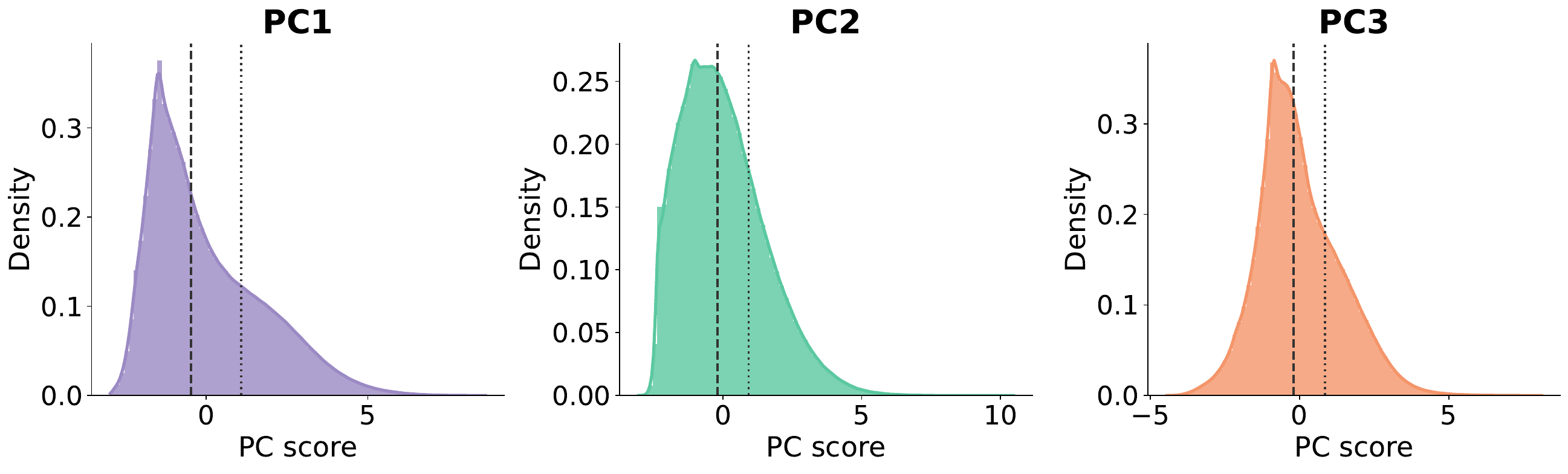}
    \caption{Distributions of the first 3 PCs across all 3.7M documents in \textsc{NarraDolma}.}
    \label{fig:pca_score_dist}
\end{figure}

\begin{figure}[th]
    \centering
    \includegraphics[width=1\linewidth]{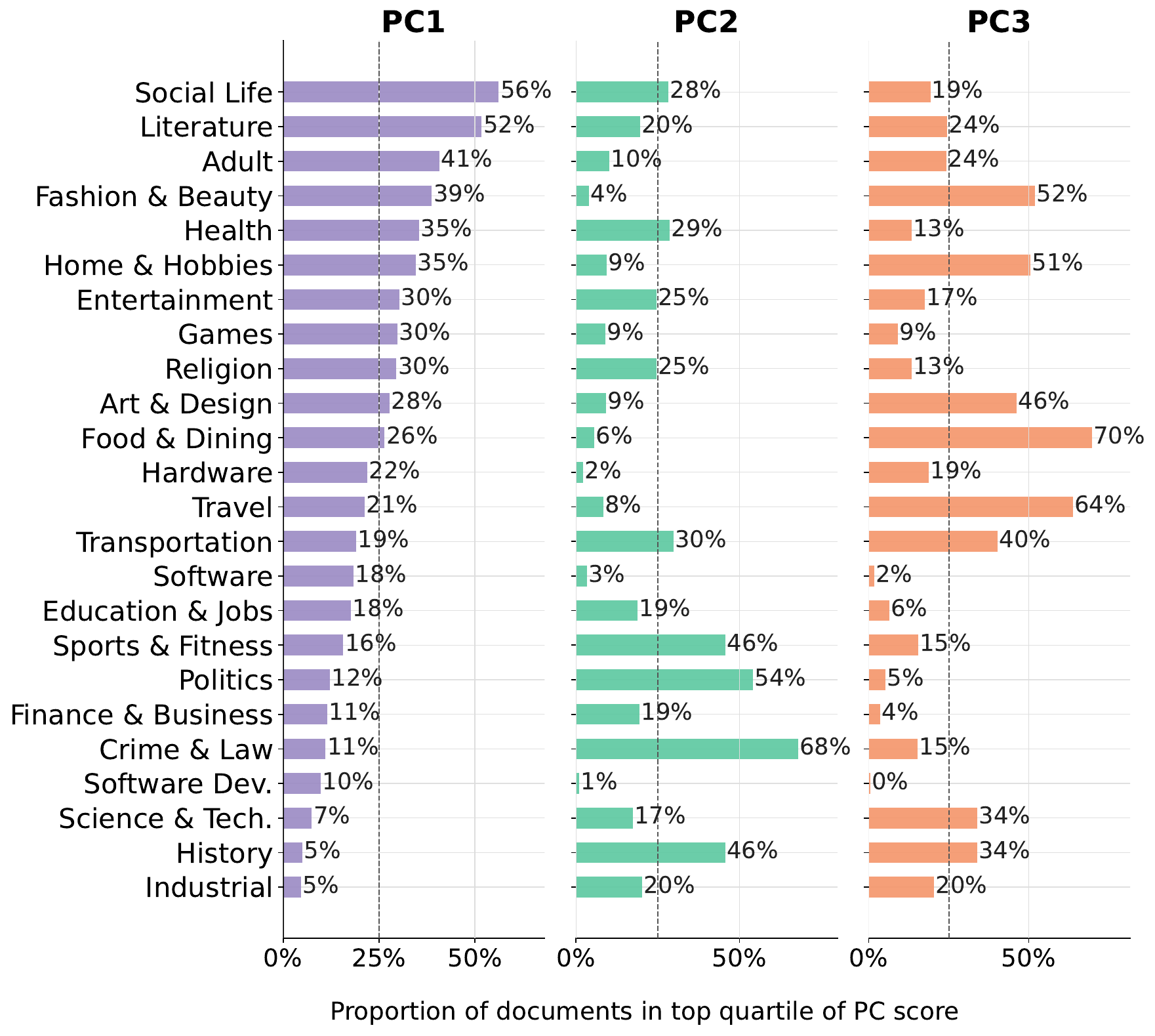}
    \caption{Bar plot of the proportion of documents within \textsc{NarraDolma} that lie within the top quartile of PC scores for each of the first three PCs}
    \label{fig:pca_topic_quartiles}
\end{figure}

Figure~\ref{fig:pca_score_dist} shows the distribution of PC scores over the full \textsc{NarraDolma} corpus (with 10 features, removing the two event relation features). Figure~\ref{fig:pca_topic_quartiles} shows a bar plot of the proportion of documents within \textsc{NarraDolma} that lie within the top quartile of PC scores for each of the first three PCs. Table~\ref{tab:pc_extremes} shows passages that have high PC scores for the first three PCs.

\begin{table}[th]
    \centering
    \includegraphics[width=1\linewidth]{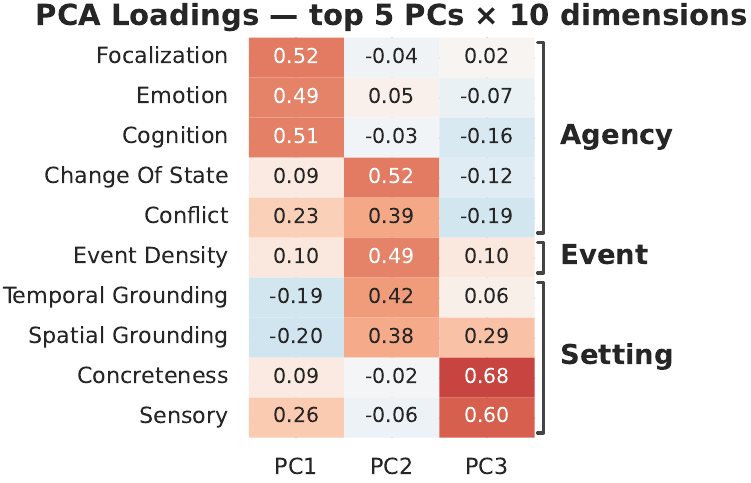}
    \caption{Loadings for the first three principal components, explaining $\sim$71\% of variance (see Fig.~\ref{fig:pc_scree}). }
    \label{tab:pc_loadings}
\end{table}

\begin{figure}[th]
    \centering
    \includegraphics[width=1\linewidth]{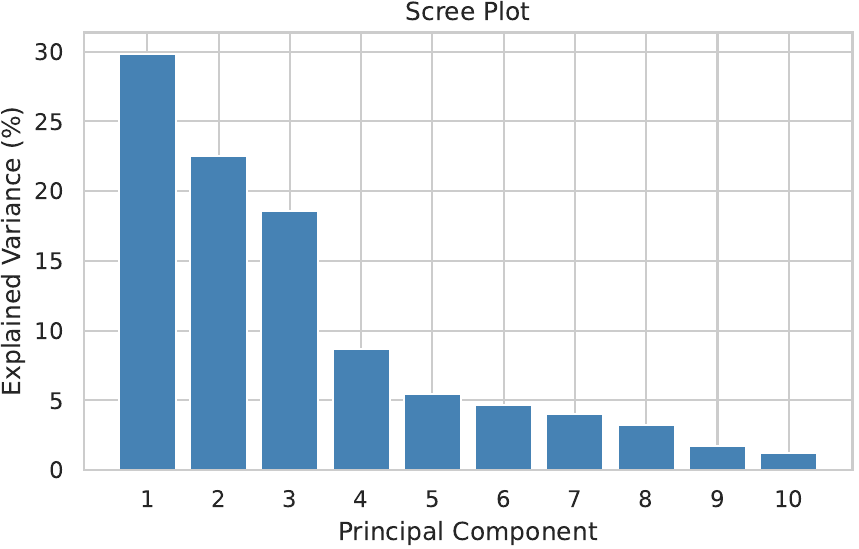}
    \caption{Scree plot of the 10 principal components across the 10 narrative features. We see a significant drop after PC3 to below 10\%.}
    \label{fig:pc_scree}
\end{figure}

\begin{table*}[th]
    \centering
    \small
    \begin{tabular}{p{1.2cm} p{0.6cm} p{1.4cm} p{11.5cm}}
    \toprule
    \textbf{PC} & \textbf{Score} & \textbf{Topic} & \textbf{Passage} \\
    \midrule
    \multirow{6}{*}{\shortstack[l]{PC1}}
    & +8.22 & Literature
    & \textit{In my head everything was a washed-out shade of grey, the color leached away by horror and utter shock. I was still there—in the café, listening to him scream and scream, staring at myself as I stood frozen, the empty cup still clutched in my hand. Wasn’t my fault.} \\
    \cmidrule{2-4}
    & $-$3.31 & History
    & \textit{University of Glasgow Research Conference, Glasgow, UK, 16-17 June 2021. Struan, A. and Alexander, M. (2010) Expressions of Civilisation and Colonisation in the Historical Thesaurus of the Oxford English Dictionary. From the Grand Tour to Mass Tourism: The Modern History of the British Abroad, Newcastle, UK, 01-02 Apr 2010.} \\
    \midrule
    \multirow{6}{*}{\shortstack[l]{PC2}}
    & +7.92 & Crime \& Law
    & \textit{But at Norwich Magistrates Court on Thursday (August 25) she was found guilty after a doorman who witnessed the attack described seeing her punch the other woman in the face to leave her bloodied. Graham Howton said he had seen her demanding to know who had her handbag before throwing the punch.} \\
    \cmidrule{2-4}
    & $-$3.20 & Reddit
    & \textit{I have a few artists in mind, but I am not sure what the subject should be. I would like to have something meaningful that symbolizes growth; i don't really know of any nonreligious symbols/art/designs that would be good for that.} \\
    \midrule
    \multirow{6}{*}{\shortstack[l]{PC3}}
    & +6.14 & Literature
    & \textit{The sun peeks out long enough to highlight the changing hues, but grayish clouds are skittering across the sky, pushed by a blustery wind that is quickly relieving all the ash trees of their gorgeous golden leaves. As I walk to the post office, my hair blows in my face and the leaves swirl and dance into piles by the sidewalk.} \\
    \cmidrule{2-4}
    & $-$5.00 & Reddit
    & \textit{There was a lot of abuse and violence towards me during my early years and I've had to deal with the ramifications, but in order for me to heal I needed to forgive her for what she did, which ultimately means letting the mother figure I craved in my life die.} \\    
    \bottomrule
    \end{tabular}
    \caption{Representative passages at the positive and negative extremes of each principal component. High-scoring passages exemplify the narrative quality captured by each axis while low-scoring passages illustrate its absence.}
    \label{tab:pc_extremes}
\end{table*}

\clearpage

\subsection{Narrative Feature Distribution}
\label{sec:feature_var}
\begin{figure}[th]
    \centering
    \includegraphics[width=1\linewidth]{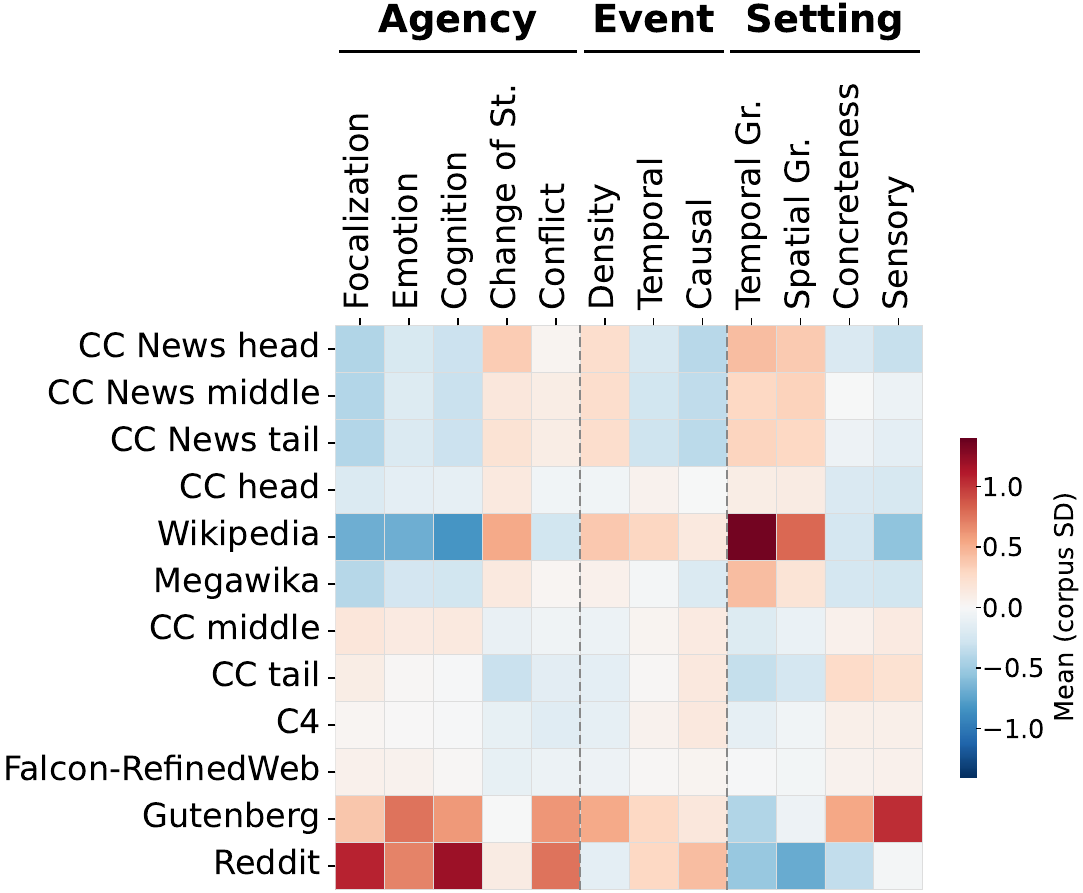}
    \caption{\textbf{Mean} $z$-scored narrative features by \textbf{\textsc{Dolma} source}, ordered by hierarchical clustering on profile similarity.}
    \label{fig:subcorpus_mean_heatmap}
\end{figure}

\begin{figure}[th]
    \centering
    \includegraphics[width=1\linewidth]{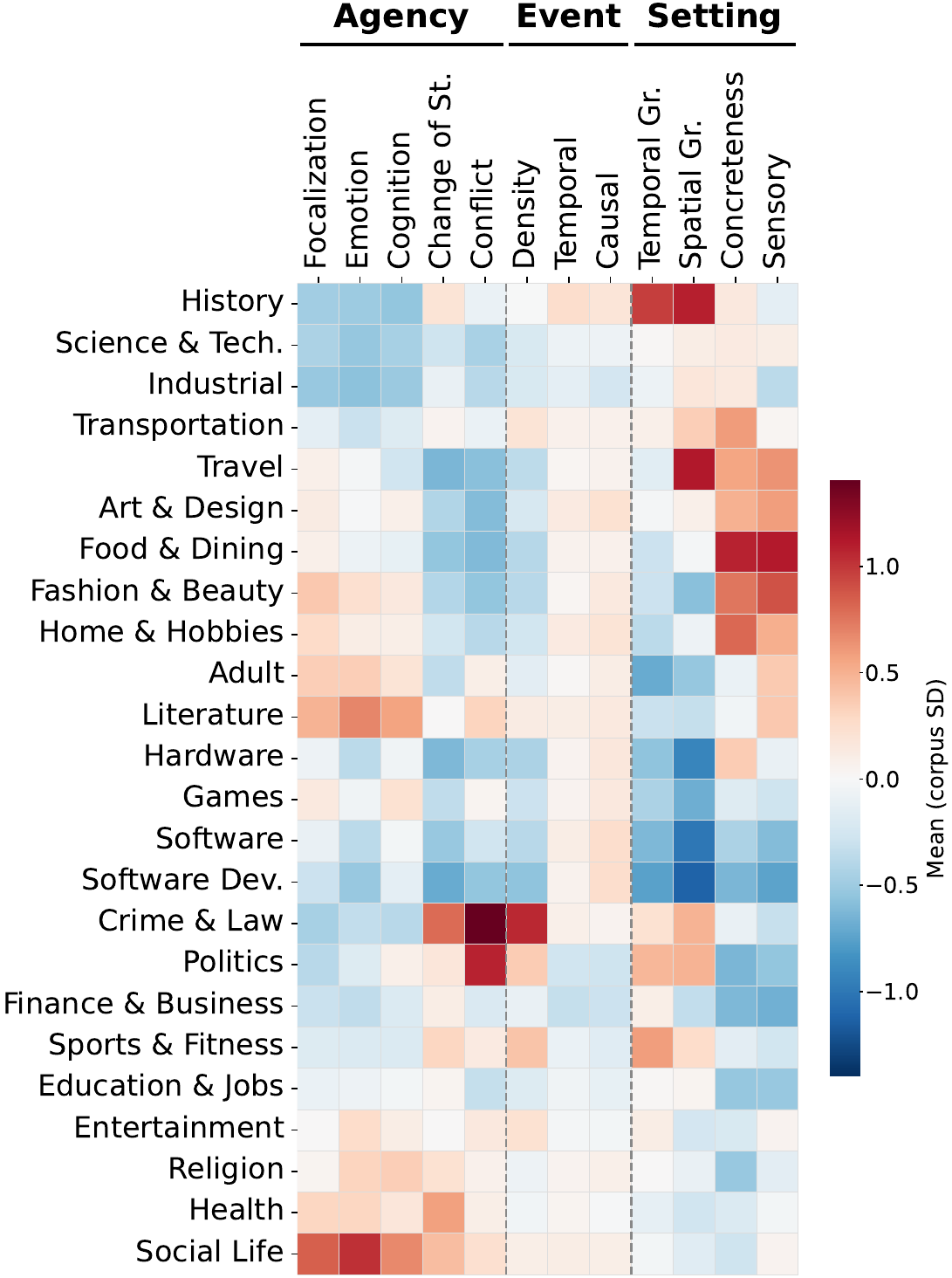}
    \caption{\textbf{Mean} $z$-scored narrative features by \textbf{topic}, ordered by hierarchical clustering on profile similarity.}
    \label{fig:topic_mean_heatmap}
\end{figure}

\begin{figure}[th]
    \centering
    \includegraphics[width=1\linewidth]{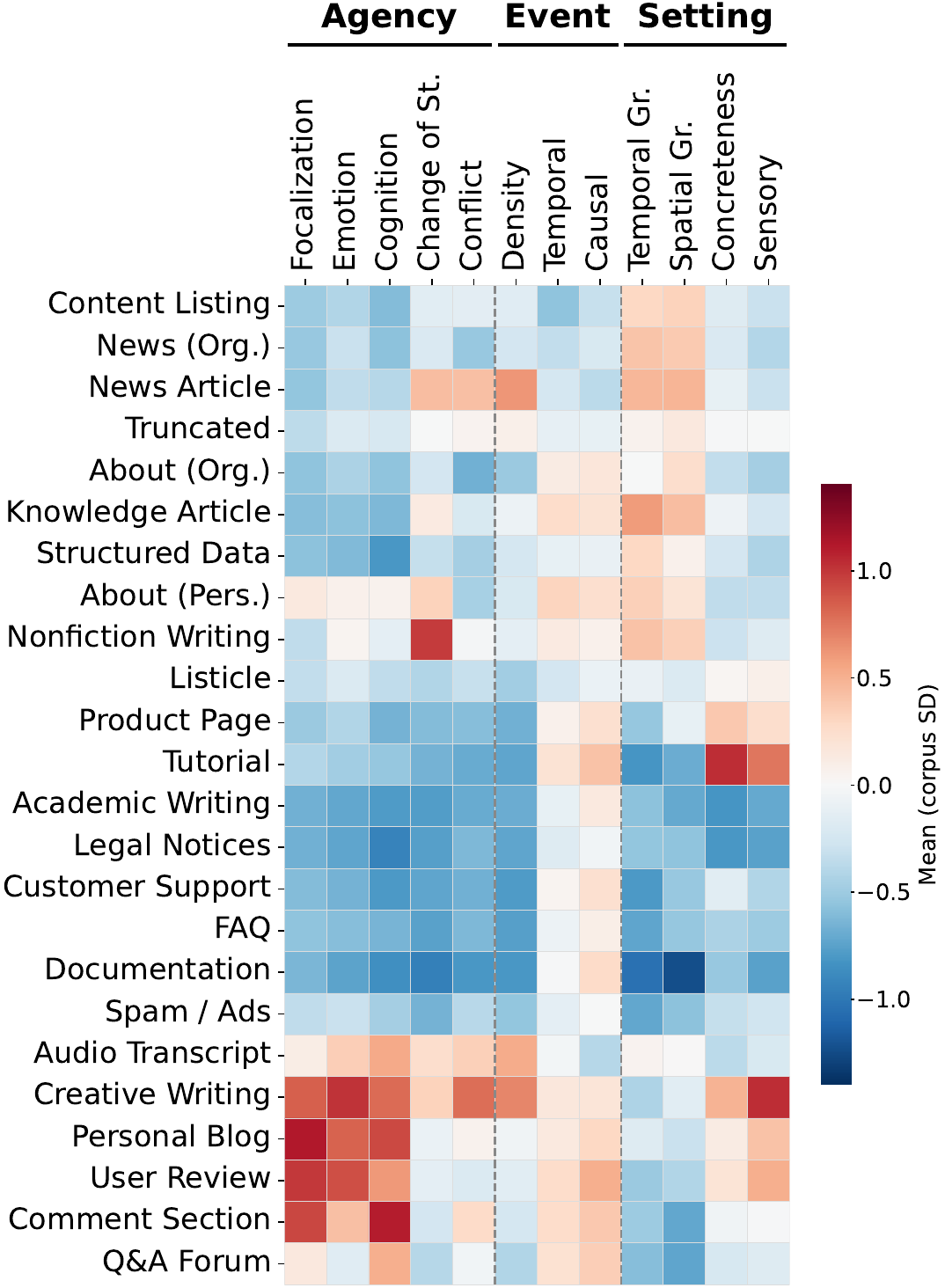}
    \caption{\textbf{Mean} $z$-scored narrative features by \textbf{format}, ordered by hierarchical clustering on profile similarity.}
    \label{fig:format_mean_heatmap}
\end{figure}

This section contains additional plots used for the analysis. Figs.~\ref{fig:subcorpus_mean_heatmap}, \ref{fig:topic_mean_heatmap}, and \ref{fig:format_mean_heatmap} show the mean z-scored narrative features. 

\clearpage

\begin{figure}[th]
    \centering
    \includegraphics[width=1\linewidth]{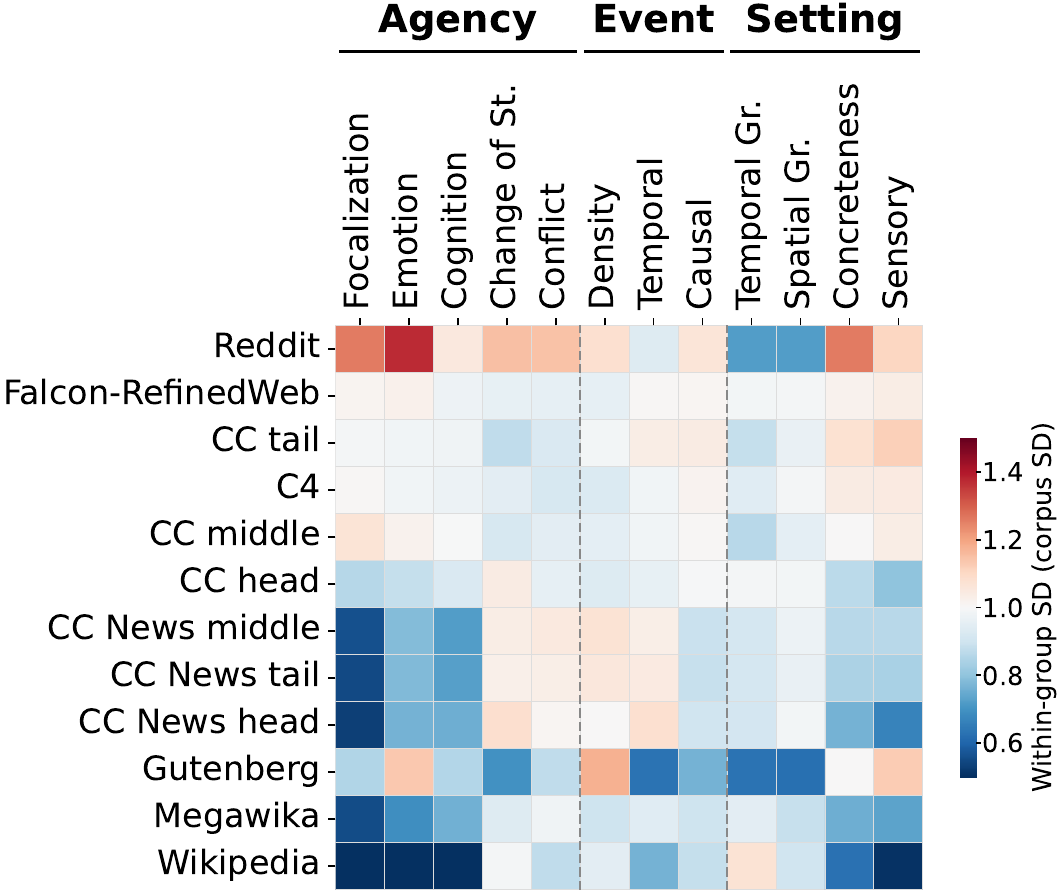}
    \caption{\textbf{SD} of $z$-scored narrative features by \textbf{source}, ordered by hierarchical clustering on profile similarity.}
    \label{fig:subcorpus_sd_heatmap}
\end{figure}

\begin{figure}[th]
    \centering
    \includegraphics[width=1\linewidth]{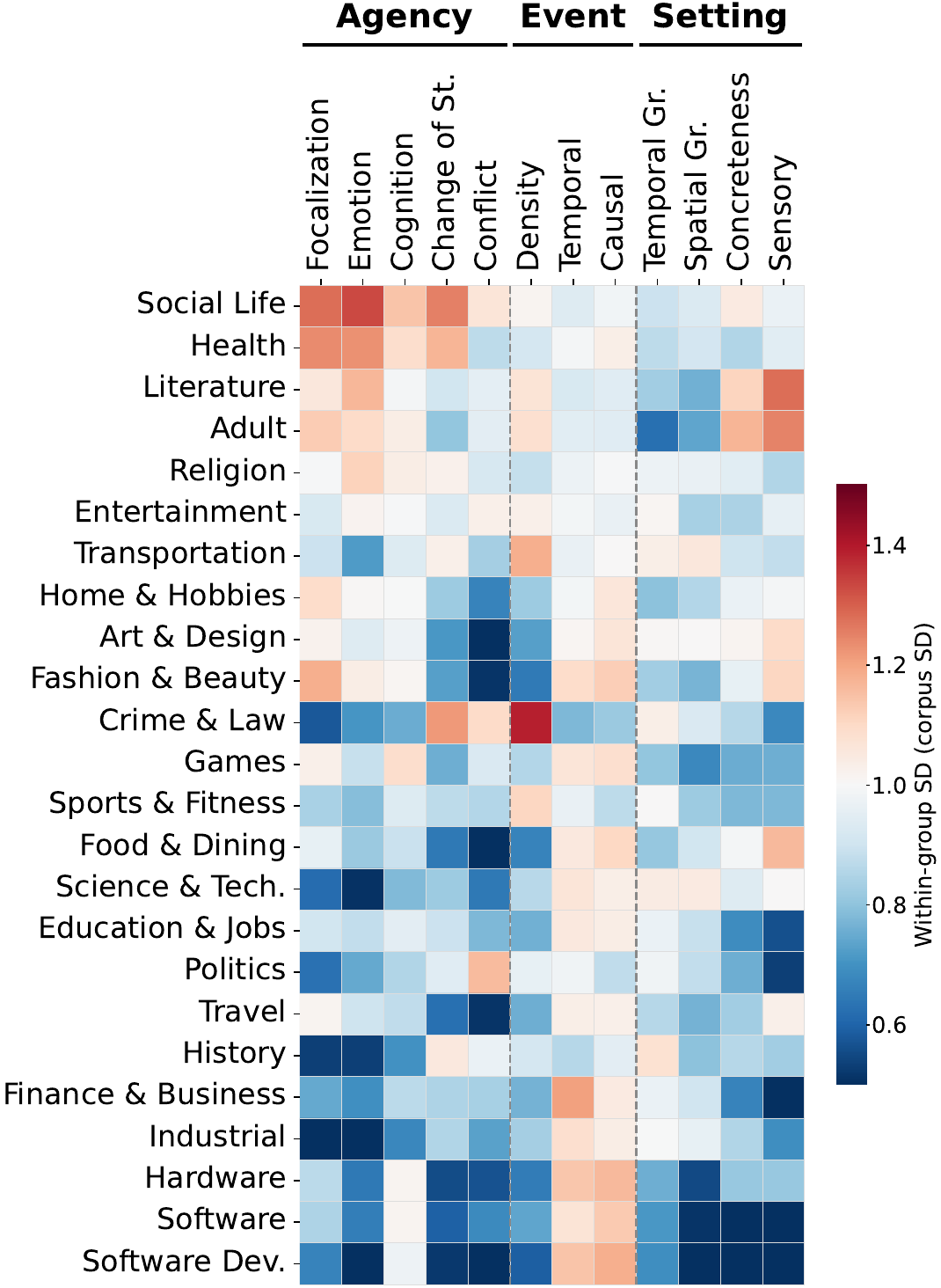}
    \caption{\textbf{SD} of $z$-scored narrative features by \textbf{topic}, ordered by hierarchical clustering on profile similarity.}
    \label{fig:topic_sd_heatmap}
\end{figure}

\begin{figure}[th]
    \centering
    \includegraphics[width=1\linewidth]{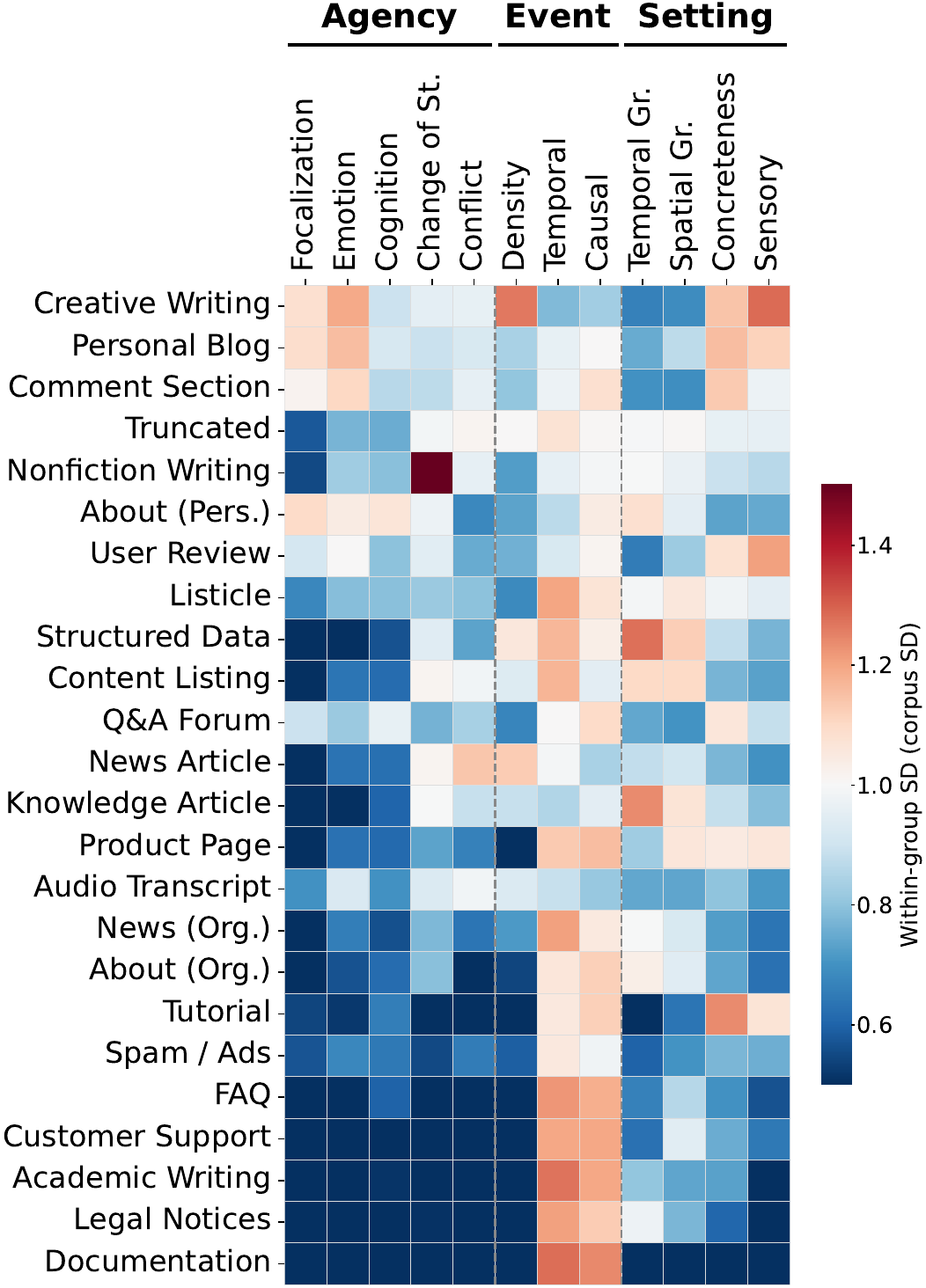}
    \caption{\textbf{SD} of $z$-scored narrative features by \textbf{format}, ordered by hierarchical clustering on profile similarity.}
    \label{fig:format_sd_heatmap}
\end{figure}

Figs.~\ref{fig:subcorpus_sd_heatmap}, \ref{fig:topic_sd_heatmap}, and \ref{fig:format_sd_heatmap} show the standard deviations (SDs) of the z-scored narrative features.

\clearpage
\section{Sampling \& Pipeline}
This appendix section details how passages were drawn from \textsc{Dolma}: the source and topic distributions of the sample, the sources we excluded and why, the full sampling configuration, and the validation of the event-span detector that underlies our event features. The final sample makes up the \narradolma corpus.
\clearpage
\subsection{Dolma Source Distributions}
\label{sec:source_dist_sec}

\begin{figure}[th]
    \centering
    \includegraphics[width=1\linewidth]{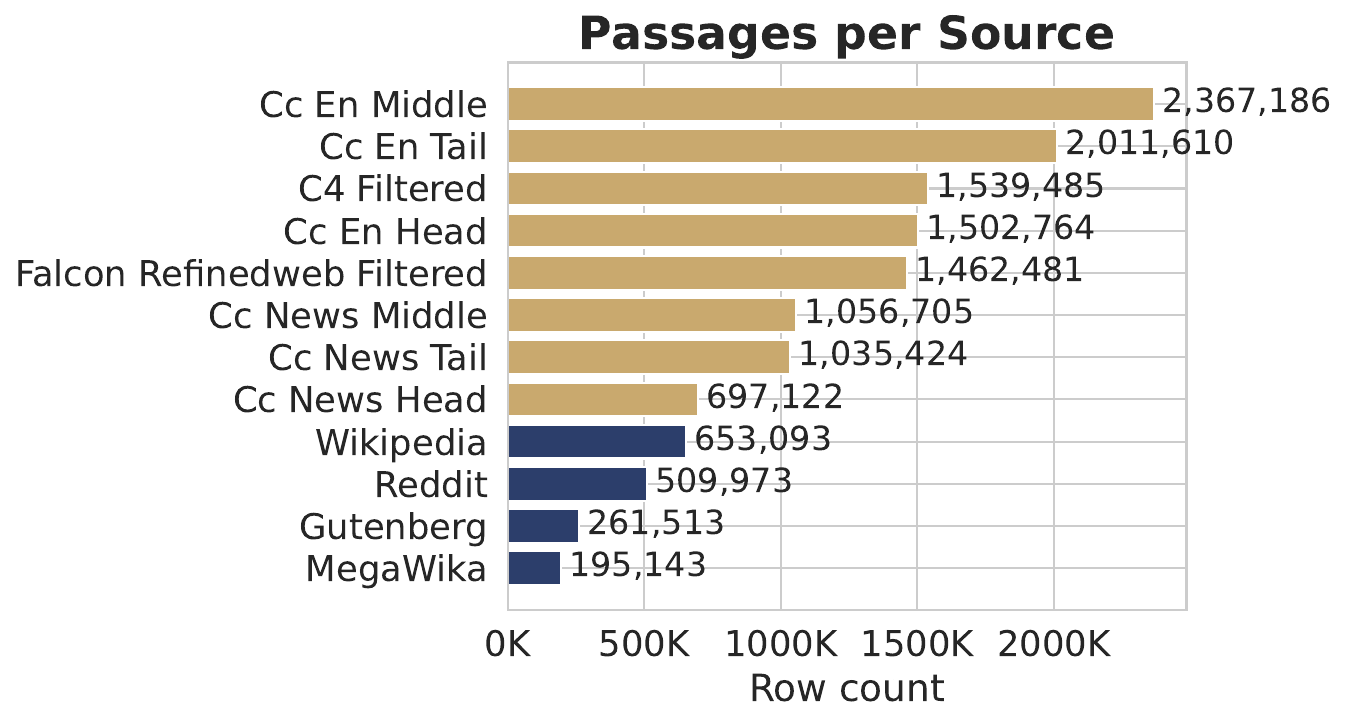}
    \caption{Distribution of Dolma sources that we sampled for \textsc{NarraDolma}. These values are at the level of passages.}
    \label{fig:narradolma_source_distribution}
\end{figure}

\begin{figure}[th]
    \centering
    \includegraphics[width=1\linewidth]{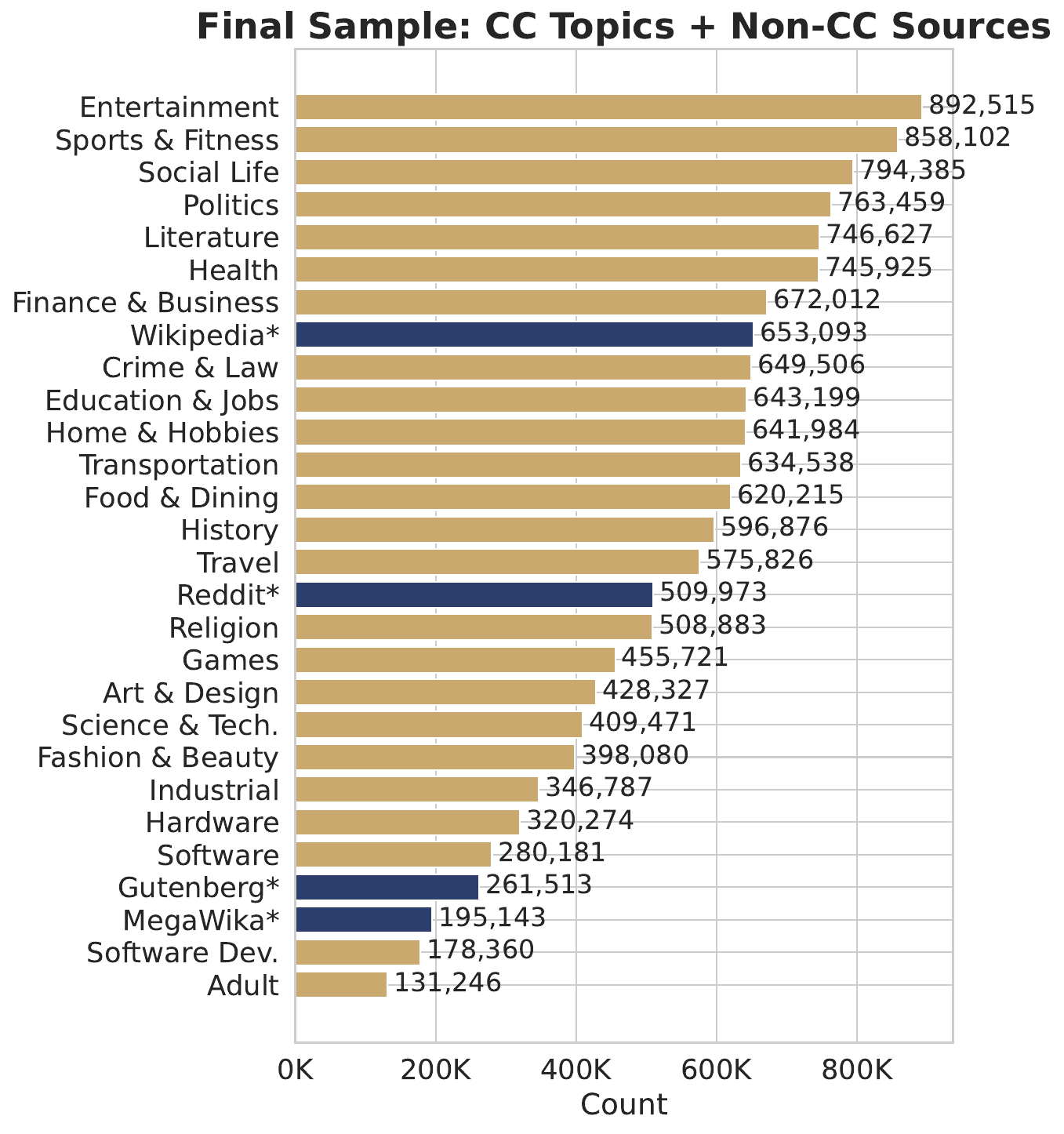}
    \caption{Distribution of Common Crawl topics and non-CC Dolma sources. Tan colors are the aggregated Common Crawl-related Dolma sources partitioned by topic from the WebOrganizer topic classifications. These values are at the level of passages.}
    \label{fig:narradolma_combined_topic_source_dist}
\end{figure}

Figures~\ref{fig:narradolma_source_distribution} and \ref{fig:narradolma_combined_topic_source_dist} show the distributions of Dolma sources and expanded categories, respectively. Table~\ref{tab:sources} shows the distribution of the gold set and \textsc{NarraDolma} corpus.

\begin{table}[th]
    \centering
    \small
    \begin{tabular}{lrrr}
    \toprule
    \textbf{Source} & \textbf{Target} & \textbf{Gold} & \textbf{\textsc{NarraDolma}} \\
    \midrule
    CC head            & 10\% &  49 &  1,502,764 \\
    CC middle          & 16\% &  72 &  2,367,186 \\
    CC tail            & 16\% &  66 &  2,011,610 \\
    C4 (filtered)      & 11\% &  35 &  1,539,485 \\
    Falcon-RefinedWeb  & 10\% &  42 &  1,462,481 \\
    \addlinespace
    CC News head       & 5\% &   14 &   697,122 \\
    CC News middle     & 8\% &   27 &   1,056,705 \\
    CC News tail       & 8\% &   27 &   1,035,424 \\
    \addlinespace
    Wikipedia          & 4\% &   8 &   653,093 \\
    Megawika           & 2\% &   4 &   195,143 \\
    \addlinespace
    Project Gutenberg  & 5\% &   21 &   261,513 \\
    \addlinespace
    Reddit             & 5\% &   35 &   509,973 \\
    \midrule
    \textbf{Total}     & 100\% & 400 & 13,292,499 \\
    \bottomrule
    \end{tabular}
    \caption{Number of 3-sentence passages sampled across \textsc{Dolma} sources. \textit{Target} is the normalized sampling proportion, the same proportions used in \textsc{Dolma}. \textit{Gold}\ is the per-source allocation for the human-annotated sample. \textit{\textsc{NarraDolma}} is the allocation for the full automatically annotated corpus.}
    \label{tab:sources}
\end{table}
\subsection{Excluded Sources from Dolma}

Table~\ref{tab:excluded} lists Dolma sources that were explicitly excluded (sampling weight set to zero) and the rationale for each.

\begin{table}[th]
    \centering
    \small
    \begin{tabular}{ll}
    \toprule
    \textbf{Source} & \textbf{Reason for exclusion} \\
    \midrule
    Proof Pile~2 (Algebraic Stack) & Formal mathematics  \\
    Proof Pile~2 (OpenWebMath)     & Formal mathematics  \\
    StarCoder                      & Source code         \\
    RedPajama-arXiv                & Academic abstracts  \\
    RedPajama-Stack Exchange       & Q\&A, code snippets \\
    PeS2o                          & Scientific papers   \\
    FLAN-TULU                      & Instruction templates \\
    \bottomrule
    \end{tabular}
    \caption{Dolma sources excluded from sampling.}
    \label{tab:excluded}
\end{table}

\clearpage

\subsection{Hyperparameters}
\label{sec:hyperparams}

Table~\ref{tab:hparams} lists all configuration parameters used across the pipeline steps.

\begin{table}[th]
    \centering
    \small
    \begin{tabular}{ll}
    \toprule
    \textbf{Parameter} & \textbf{Value} \\
    \midrule
    Initial pool size              & 51M passages\\
    Unique documents               & 15M documents\\
    Chunk length                   & 3 sentences \\
    \addlinespace
    Narrative classifier           & RoBERTa \\
    \quad batch size               & 64   \\
    \quad max input length         & 512 tokens   \\
    \quad Narrative confidence threshold & $p \geq 0.50$ \\
    Topic classifier               & WebOrganizer \\
    \quad batch size               & 32 \\
    \quad max input length         & 512 tokens \\
    \addlinespace
    LLM model for annotators                     & Llama-3.1-8B-Instruct \\
    \quad quantization             & 4-bit NF4 (bfloat16) \\
    \quad snippet length           & 1{,}200 chars $\times$ \\
    \quad max new tokens           & 120  \\
    \quad temperature              & 0.2 \\
    \quad top-$p$                  & 0.9 \\
    \quad summary character cap    & 300 chars \\
    \bottomrule
    \end{tabular}
    \caption{Full configuration for the Dolma sampling pipeline.}
    \label{tab:hparams}
\end{table}

\paragraph{RoBERTa Training.} We fine-tune roberta-base with nine task-specific linear regression heads, one per narrative dimension, appended to the [CLS] token representation. Training uses 5-fold cross-validation on the gold-annotated instances. Within each fold we optimize with AdamW (learning rate $2 \times 10^{-5}$, weight decay $0.01$) using a linear warmup schedule over the first 10\% of training steps, a batch size of 16, and a maximum sequence length of 200 tokens. Training runs for up to 20 epochs with early stopping (patience = 3) monitored by validation MAE on the held-out fold; the final model is then retrained on all gold instances for the median number of best epochs across folds. The loss is masked MSE, which ignores any NaN-valued labels in the multi-task objective. Gradients are clipped to a maximum norm of 1.0. All experiments use random seed 42. Training was performed on an NVIDIA H100 GPU (40 GB MIG partition) on the CU Boulder Blanca research computing cluster. \textsc{RoBERTa-base} is roughly 125M parameters, the 9 regression heads add $\sim$7K.

\paragraph{RoBERTa Inference.} Inference over the gold evaluation set uses a batch size of 64 with the same tokenizer settings as training (max length 200, padding to max length, truncation). Raw regression outputs are clipped to the valid Likert range [1, 5] before computing metrics.

\paragraph{LLM Inference.} All three LLMs receive identical prompts: the same system prompt and user message are used across models, with scores elicited as integers on a 1–5 scale. No temperature or sampling parameters are explicitly set; each model uses its provider default.
\begin{itemize}
    \item Claude Sonnet 4.6 is called via the Anthropic Message Batches API with forced tool use (tool\_choice: annotate\_narrative, max\_tokens: 512). The system prompt and tool definition are cached with ephemeral prompt caching to reduce latency and cost.
    \item Gemma 4 31B and Qwen3-235B-A22B are called via the Doubleword AI async responses API (service\_tier: flex, background: true) with 20 concurrent workers. Scores are parsed from the model's plain-text JSON output, with a regex fallback extractor for malformed responses. \textsc{Gemma 4 31B} is roughly 31B parameters and \textsc{Qwen3-235B-A22B} is roughly 235B total parameters (22B active).
\end{itemize}
\clearpage
\subsection{Event Span Detection}
\label{sec:event_span_detection}

Event trigger spans are the lexical anchors around which event meaning is organized \cite{sims-etal-2019-literary}. We detect two complementary types of spans in each passage in order to verify the use of the LitBank event detector for our dataset:

\paragraph{Event spans.}
Event spans are identified using a RoBERTa-based event detector trained on LitBank \cite{sims-etal-2019-literary}, a corpus of literary text annotated with event mentions following ACE event type definitions. On average, each passage contains 2.4 event spans.

\paragraph{Verb spans.}
We additionally extract verb spans using the \textsc{spaCy} \textsc{en\_core\_web\_trf} pipeline\footnote{https://spacy.io/models/en}. On average, each passage contains 6.9 verb spans. Verb spans that overlap with any event span are discarded to avoid redundancy.

\paragraph{Span Pair Selection.}
From each passage's combined pool of event and verb spans, we select exactly one adjacent pair for manual annotation. Ultimately we find that the LitBank event detection achieves an F1 of 85\%, with a precision of 90\%. We move forward with the LitBank event detection model as our event detector.

\section{Multilingual Distribution Comparison}
\label{sec:multilingual_dist}

Our work is conducted over the English language. To explore potential multilingual capabilities of our model, we compare the distribution of scores of a sample of Common Crawl web text across English, Arabic, German, and French. Samples are drawn from FineTranslations \cite{penedo2026finetranslations}, a release of FineWeb2 machine-translated into English. This dataset contains text that has already been translated from each of these languages to English, so our model is given just English text.

\begin{figure*}
    \centering
    \includegraphics[width=1\linewidth]{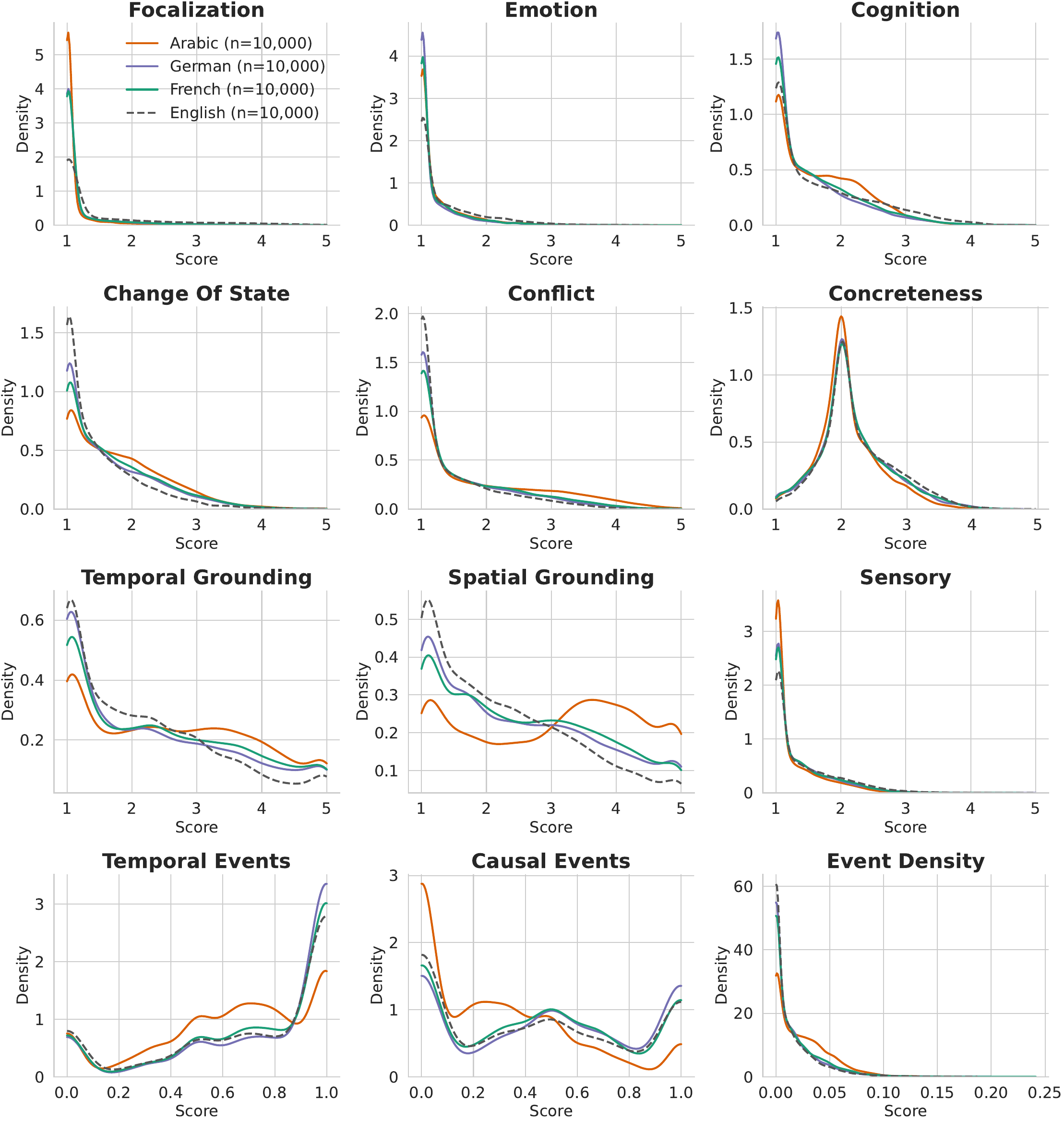}
    \caption{Distribution of the 10K documents' narrative scores across English, Arabic, German, and French.}
    \label{fig:multilingual_dist}
\end{figure*}

For each language we sample 10,000 documents, which are then sentence-tokenized and cut into 3-sentence chunks. We then keep up to three passages per document which mimicks our Dolma sampling procedure. We apply the topic, narrative classifier, and event detection models to each passage. Fig.~\ref{fig:multilingual_dist} shows the distributions of each language across all 12 narrative features. In general, German and French are closest to English across all features. We do notice that the translated Arabic texts have higher spatial grounding.

\clearpage
\section{Annotation Framework}
This appendix documents the full annotation framework: the rating scale for each of the 11 dimensions, the interface annotators used, the document summaries shown to orient them, and the prompts used to elicit the LLM annotations.
\subsection{LLM-Generated Document Summaries for Human Annotation}
\label{sec:llmsummary}

To orient annotators to the genre and topic of each passage without biasing their event, agency, and setting judgments, we pair each chunk with a short machine-generated description of the originating document. We run Llama-3.1-8B-Instruct \cite{llama3} in 4-bit NF4 quantization over each retrieved document, providing the model with three 1,200-character snippets drawn from the start, middle, and end of the full text.

The system message provided to Llama-3.1-8B-Instruct for summary
generation is:

\begin{quote}
    \small
    \textit{``You write short, extremely surface-level summaries of webpages. Do not speculate or interpret. Only describe what is explicitly present.''}
\end{quote}

\subsection{Potato Interface}
\label{sec:potato_ui}

\begin{figure*}[t]
    \centering
    \includegraphics[width=1\linewidth]{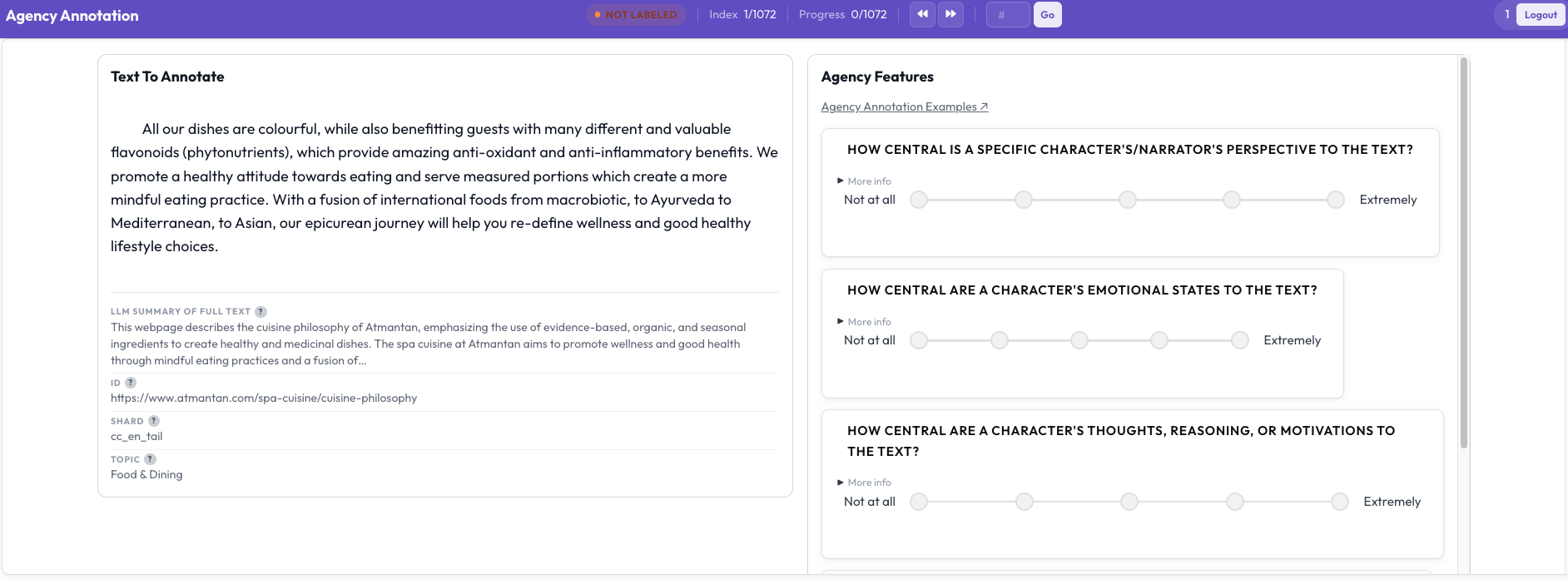}
    \caption{Screenshot of the agency annotation UI built upon the Potato software.}
    \label{fig:agency_potato_screenshot}
\end{figure*}

\begin{figure*}[t]
    \centering
    \includegraphics[width=1\linewidth]{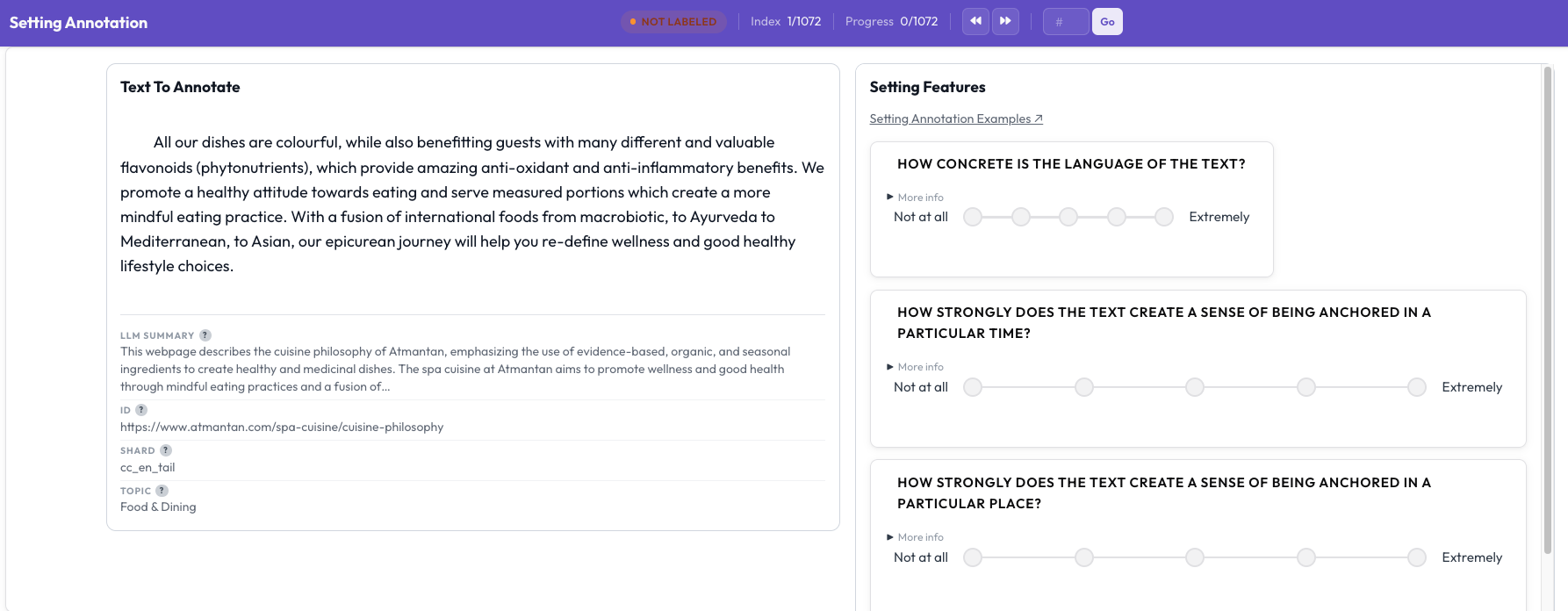}
    \caption{Screenshot of the setting annotation UI built upon the Potato software.}
    \label{fig:setting_potato_screenshot}
\end{figure*}

\begin{figure*}[t]
    \centering
    \includegraphics[width=1\linewidth]{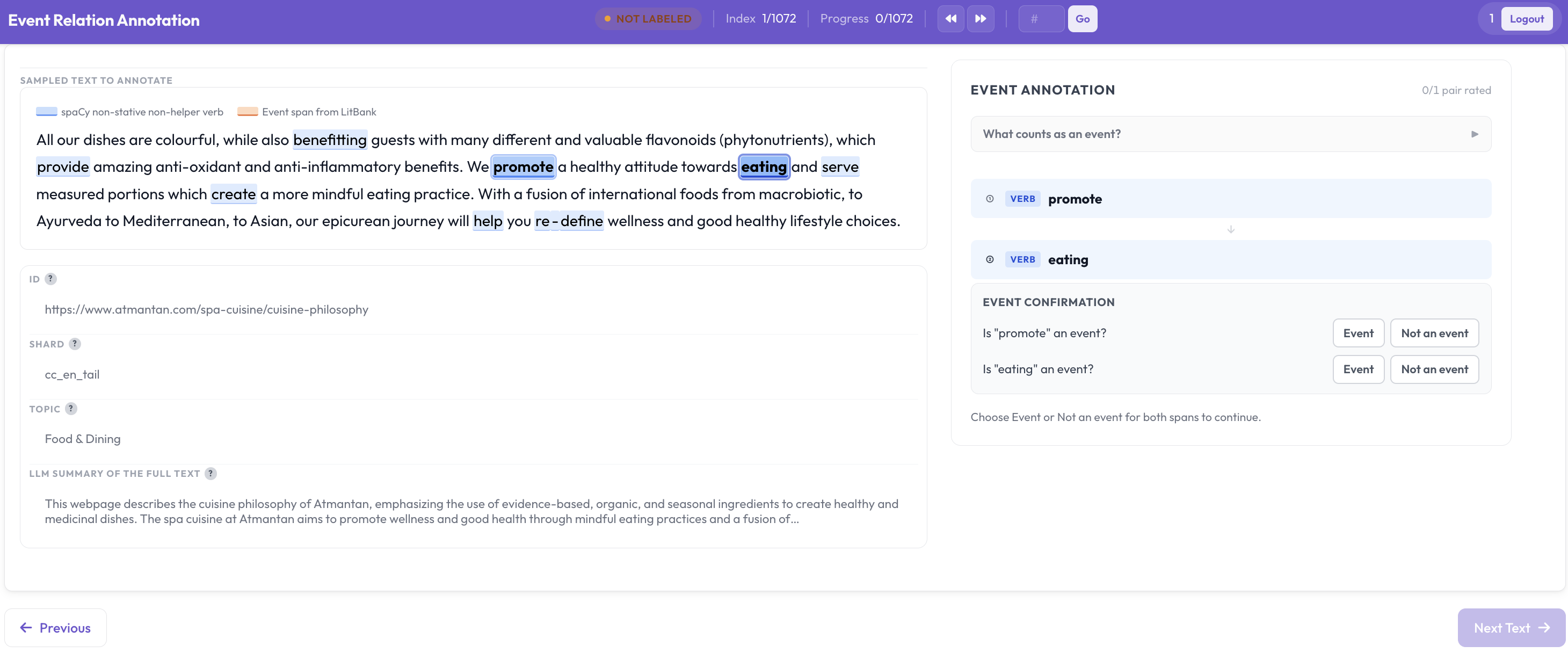}
    \caption{Screenshot of the event relation annotation UI built upon the Potato software.}
    \label{fig:event_potato_screenshot}
\end{figure*}

Figures~\ref{fig:agency_potato_screenshot}, \ref{fig:setting_potato_screenshot}, and \ref{fig:event_potato_screenshot} show the UI that annotators use to label the sampled texts for agency and setting, respectively.

\clearpage
\subsection{Annotator Feature scales}
\label{sec:feature_scales}

Tables~\ref{tab:focalization}-\ref{tab:sensory} show the feature scales given to annotators for each of the 11 fine-grained narrative features.

\begin{table}[th]
    \centering
    \small
    \begin{tabular}{@{}p{0.6cm}p{6.5cm}@{}}
    \toprule
    \textbf{Score} & \textbf{Description} \\
    \midrule
    1 & Not central. The passage describes events and details from a purely external vantage point with no access to any character's inner life. \\
    \addlinespace
    2 & Minimally central. A character's perspective is faintly implied but does not shape how events are presented. The passage remains largely external. \\
    \addlinespace
    3 & Moderately central. A character's perspective is present and shapes some of the passage, but external description and internal access are roughly balanced. \\
    \addlinespace
    4 & Considerably central. Events are predominantly filtered through a character's perceptions and inner life, with only occasional external description. \\
    \addlinespace
    5 & Extremely central. The passage is fully organized around a specific character's or narrator's perspective. Every detail is filtered through their perceptions, feelings, and inner experience. \\
    \bottomrule
    \end{tabular}
    \caption{Focalization rating scale.}
    \label{tab:focalization}
\end{table}

\begin{table}[th]
    \centering
    \small
    \begin{tabular}{@{}p{0.6cm}p{6.5cm}@{}}
    \toprule
    \textbf{Score} & \textbf{Description} \\
    \midrule
    1 & Not central. The passage contains no reference to how a character feels, either through observable behavior or internal access. \\
    \addlinespace
    2 & Minimally central. A character's emotional state is faintly implied through observable behavior (e.g., a gesture or expression), but feelings are not elaborated or sustained. \\
    \addlinespace
    3 & Moderately central. A character's emotional states are present and noticeable, but emotion is not the dominant or organizing feature of the passage. \\
    \addlinespace
    4 & Considerably central. Emotional experience is prominent and shapes how the passage is organized. The reader has meaningful access to how a character feels, either through behavior or internal description. \\
    \addlinespace
    5 & Extremely central. Emotional experience dominates the passage. The reader is given sustained and direct access to a character's feelings, which are foregrounded throughout. \\
    \bottomrule
    \end{tabular}
    \caption{Internal emotion rating scale.}
    \label{tab:emotion}
\end{table}

\begin{table}[th]
    \centering
    \small
    \begin{tabular}{@{}p{0.6cm}p{6.5cm}@{}}
    \toprule
    \textbf{Score} & \textbf{Description} \\
    \midrule
    1 & Not central. The passage contains no reference to what a character thinks, believes, reasons, or wants. \\
    \addlinespace
    2 & Minimally central. A character's mental state is faintly implied but not elaborated. Cognition is incidental to the passage. \\
    \addlinespace
    3 & Moderately central. A character's thoughts or reasoning are present and contribute to the passage, but are not its dominant or organizing feature. \\
    \addlinespace
    4 & Considerably central. A character's beliefs, deliberations, goals, or reasoning are prominent and shape how the passage is organized. \\
    \addlinespace
    5 & Extremely central. The passage is dominated by a character's inner cognitive life including their thoughts, reasoning, beliefs, and desires. \\
    \bottomrule
    \end{tabular}
    \caption{Internal cognition rating scale.}
    \label{tab:cognition}
\end{table}
 
\begin{table}[th]
    \centering
    \small
    \begin{tabular}{@{}p{0.6cm}p{6.5cm}@{}}
    \toprule
    \textbf{Score} & \textbf{Description} \\
    \midrule
    1 & Not central. Characters remain essentially unchanged across the passage. No change in physical, psychological, relational, or existential condition is present or implied. \\
    \addlinespace
    2 & Minimally central. A minor or incidental change is present or implied, but it is not an organizing feature of the passage. \\
    \addlinespace
    3 & Moderately central. A change in a character's condition is present and contributes meaningfully to the passage, but is not its dominant feature. \\
    \addlinespace
    4 & Considerably central. A change in a character's condition -- physical, psychological, relational, or existential -- is prominent and shapes how the passage is organized. \\
    \addlinespace
    5 & Extremely central. Change is the dominant and organizing feature of the passage. The transformation (whether completed, in progress, or strongly implied) is what the passage is fundamentally about. \\
    \bottomrule
    \end{tabular}
    \caption{Change of state rating scale.}
    \label{tab:changeofstate}
\end{table}

\begin{table}[th]
    \centering
    \small
    \begin{tabular}{@{}p{0.6cm}p{6.5cm}@{}}
    \toprule
    \textbf{Score} & \textbf{Description} \\
    \midrule
    1 & Not central. No conflict of any kind is present or implied. The passage describes actions or states without opposition or tension. \\
    \addlinespace
    2 & Minimally central. A hint of tension or opposition is present but remains entirely in the background and does not shape the passage. \\
    \addlinespace
    3 & Moderately central. Conflict is present and noticeable -- between characters, within a character, or against external forces -- but is not the dominant feature of the passage. \\
    \addlinespace
    4 & Considerably central. Conflict is prominent and shapes how the passage is organized, whether interpersonal, internal, or against external forces or circumstances. \\
    \addlinespace
    5 & Extremely central. Conflict dominates the passage. The opposition or tension is what the passage is fundamentally about, and it is sustained throughout. \\
    \bottomrule
    \end{tabular}
    \caption{Conflict rating scale.}
    \label{tab:conflict}
\end{table}

\begin{table}[th]
    \centering
    \small
    \begin{tabular}{@{}p{0.6cm}p{6.5cm}@{}}
    \toprule
    \textbf{Score} & \textbf{Description} \\
    \midrule
    1 & No physical referents. The passage is composed entirely of abstract ideas, generalizations, categories, or institutional descriptions. Nothing could be explained by pointing. \\
    \addlinespace
    2 & Physical objects or entities are named but not rendered. The text identifies things that exist in the world but provides no perceptual detail. No shape, color, texture, or material quality. \\
    \addlinespace
    3 & Physical referents are present with some rendering. Objects and actions are described with enough specificity to suggest a physical scene, but perceptual detail is sparse or thin. \\
    \addlinespace
    4 & Physical referents are rendered with clear perceptual detail. The passage provides enough specific, tangible description (through surface qualities, physical components, or rendered actions) that the reader can picture the scene. \\
    \addlinespace
    5 & Physical referents are richly and fully rendered. The passage is dense with perceptual particulars. Objects, environments, and actions are described with enough sensory and material detail that the world feels concretely present. \\
    \bottomrule
    \end{tabular}
    \caption{Concreteness rating scale.}
    \label{tab:concreteness}
\end{table}

\begin{table}[th]
    \centering
    \small
    \begin{tabular}{@{}p{0.6cm}p{6.5cm}@{}}
    \toprule
    \textbf{Score} & \textbf{Description} \\
    \midrule
    1 & No temporal content. The passage has no markers of time whatsoever; events could be happening at any point in history or none at all. \\
    \addlinespace
    2 & Temporal language present but only sequencing events. Relative markers (e.g., \textit{before}, \textit{after}, \textit{recently}, \textit{a few years ago}) order events without locating the reader in any particular time. \\
    \addlinespace
    3 & Modest temporal location. The passage provides cyclical grounding -- a season, a time of day, a day of the week -- that gives the reader a felt sense of when without historical specificity. Most cyclical-only passages cap here. \\
    \addlinespace
    4 & Strong temporal location. The passage provides historical grounding -- a specific year, era, named event, or cultural period -- that anchors the reader in an identifiable moment. In rare cases, rich and sustained cyclical grounding can reach this level. \\
    \addlinespace
    5 & Vivid and sustained temporal immersion. Both historical and cyclical grounding work together, creating a strong and atmospheric sense of a specific moment in time. \\
    \bottomrule
    \end{tabular}
    \caption{Temporal grounding rating scale.}
    \label{tab:temporal}
\end{table}

\begin{table}[th]
    \centering
    \small
    \begin{tabular}{@{}p{0.6cm}p{6.5cm}@{}}
    \toprule
    \textbf{Score} & \textbf{Description} \\
    \midrule
    1 & No spatial content. The passage has no markers of place; events could be happening anywhere or nowhere. \\
    \addlinespace
    2 & Spatial language present but minimal. Either a bare proximate location with no rendering (e.g., \textit{she sat in the kitchen}) or a vague geographic reference (e.g., \textit{somewhere in Europe}). Places the reader no further than knowing a space exists. \\
    \addlinespace
    3 & Modest spatial location. Either a named geographic location that places the reader on the globe without evoking it (e.g., a city name), or a proximate location with some modest rendering that begins to create a felt sense of place. \\
    \addlinespace
    4 & Strong spatial location. Either geographic grounding with atmospheric texture, or a proximate environment rendered vividly enough that the reader feels present in the space. \\
    \addlinespace
    5 & Vivid and sustained spatial immersion. Both geographic and proximate grounding work together with rich rendering, creating a strong and atmospheric sense of a specific place. \\
    \bottomrule
    \end{tabular}
    \caption{Spatial grounding rating scale.}
    \label{tab:spatial}
\end{table}

\begin{table}[th]
    \centering
    \small
    \begin{tabular}{@{}p{0.6cm}p{6.5cm}@{}}
    \toprule
    \textbf{Score} & \textbf{Description} \\
    \midrule
    1 & No sensory content. Events and ideas are conveyed without any appeal to the senses; no modality is invoked. \\
    \addlinespace
    2 & Sensory content present but incidental. One or more senses are mentioned or implied, but the sensory experience is peripheral to the passage's meaning. Removing the sensory language would not substantially change the passage. \\
    \addlinespace
    3 & Sensory content present and noticeable. At least one sense modality recurs or contributes meaningfully to the passage, but sensory experience is not the organizing feature of the text. \\
    \addlinespace
    4 & Sensory experience is central. One or more senses are prominent and clearly organize how the passage is structured, even if the sensory content is not rendered in rich perceptual detail. \\
    \addlinespace
    5 & Sensory experience is dominant and richly rendered. The passage is organized around sensory experience across one or more modalities, with sustained and detailed evocation of what it feels like to perceive the described world. \\
    \bottomrule
    \end{tabular}
    \caption{Sensory detail rating scale.}
    \label{tab:sensory}
\end{table}
\clearpage
\subsection{LLM Annotation Prompt: Agency and Setting}
\label{app:llm_prompt}
 
Fig.~\ref{fig:agency_prompt} displays the prompt used to elicit agency annotations and Fig.~\ref{fig:setting_prompt} the prompt for setting annotations from Claude Sonnet~4.6, Qwen3-235B-A22B, and Gemma4-31B. The prompt was held constant across all three models and across LLM validation and large-scale annotation.
 
\begin{figure*}
\setlength{\fboxsep}{10pt}
\fbox{%
\begin{minipage}{\linewidth}
    \small
     
    You are an expert at identifying narrative qualities in web text. Your task is to read a passage and rate it on nine dimensions across two categories (Agency and Setting) each on a scale from 1 (not at all) to 5 (extremely).
     
    \vspace{1pt}
    \textbf{AGENCY DIMENSIONS}
     
    \vspace{1pt}
    These five dimensions capture how characters are represented as agents in the text. Importantly, these dimensions rate the presence and centrality of the feature in the text as written, not what can be inferred about the content.
     
    \vspace{1pt}
    \textbf{1. Focalization}
     
    How central is a specific character's/narrator's perspective to the text?
     
    A higher score reflects text in which the details and events are primarily presented through a specific character's perspective: their perceptions, thoughts, and feelings shape how the story is told. A lower score reflects text that reports details and events from the outside, describing what is observable without granting access to any character's internal experience.
     
    Focalization is about whether the reader experiences events through a specific perceiving consciousness's inner life, regardless of the formal mechanism. That can be achieved through: direct quotation (if the content renders inner experience), free indirect discourse (\textit{``she couldn't shake the feeling...''}), first person narration with genuine interiority, or close third person narration. When ``you'' appears in instructional or legal text, it should score low on focalization. However a specific second-person perspective that immerses the reader can score high.
     
    \begin{itemize}[noitemsep, leftmargin=*]
        \item Score 5: \textit{``she scanned the faces in the crowd, certain that someone was watching her''}
        \item Score 3: \textit{``He wasn't sure the meeting had gone well. He gathered his things and headed for the door.''}
        \item Score 1: \textit{``she walked through the crowd.''}
    \end{itemize}
     
    \vspace{1pt}
    \textbf{2. Emotion}
     
    How central are a character's emotional states to the text?
     
    A higher score reflects text in which a character's emotional experience is a prominent feature. A lower score reflects text with little or no reference to how a character feels.
     
    \begin{itemize}[noitemsep, leftmargin=*]
        \item Score 5: \textit{``Completely paralyzed with fear, he was all of a sudden flooded with relief when he heard her voice''}
        \item Score 3: \textit{``She was nervous about the presentation. She went into the interview room.''}
        \item Score 1: \textit{``he answered the phone.''}
    \end{itemize}
     
    \vspace{1pt}
    \textbf{3. Cognition}
     
    How central are a character's thoughts, reasoning, or motivations to the text?
     
    A higher score reflects text in which a character's thoughts, beliefs, reasoning, goals, or desires are a prominent feature. A lower score reflects text with little or no reference to what a character thinks, intends, or wants.
     
    \begin{itemize}[noitemsep, leftmargin=*]
        \item Score 5: \textit{``she kept turning the problem over in her mind, convinced there was something she had missed. Was it because they were so quiet in the meeting? Or their body language? She couldn't tell.''}
        \item Score 3: \textit{``She read the email twice. It wasn't entirely clear what they were asking for, but she thought she understood the gist. She drafted a reply.''}
        \item Score 1: \textit{``she sat at her desk''}
    \end{itemize}
     
    \vspace{1pt}
    \textbf{4. Change of State}
     
    How central is a change in a character's condition or state to the text?
     
    A higher score reflects text in which a change in a character's condition/state is a prominent or organizing feature. A lower score reflects text where characters remain essentially unchanged. The change does not need to be completed within the passage; an in-progress or partially implied change still counts. If a passage describes a world event (a company going bankrupt, an earthquake, a death) that necessarily entails a change in a character's condition, that counts toward the score even if the character's internal response is not elaborated. Change of state should be rated on presence and centrality of the changes themselves, independent of how vividly or dramatically they are rendered.
     
    \begin{itemize}[noitemsep, leftmargin=*]
        \item Score 5: \textit{``by the time she reached the door, something in her had shifted. She wasn't the same person who had walked in''}
        \item Score 3: \textit{``By the end of the conversation he felt somewhat better about the situation. It wasn't resolved, but it seemed more manageable than it had before.''}
        \item Score 1: \textit{``she sat at her desk.''}
    \end{itemize}
     
    \vspace{1pt}
    \textbf{5. Conflict}
     
    How central is conflict involving characters to the text?
     
    A higher score reflects text in which conflict is a dominant or organizing feature. A lower score reflects text in which conflict is absent or only incidentally present. Conflict can take many forms: tension between characters, internal psychological struggle, or opposition to institutions, environments, inanimate objects like technology, social forces, or physical events.
     
    \begin{itemize}[noitemsep, leftmargin=*]
        \item Score 5: \textit{``they had been arguing for hours, neither willing to give ground. The people around them started to stare. They were so caught up in the fight that they didn't even notice.''}
        \item Score 3: \textit{``They disagreed about the route. They took the highway and didn't talk much for the first hour.''}
        \item Score 1: \textit{``they sat across from each other at the table.''}
    \end{itemize}
\end{minipage}%
}
\caption{Annotation prompt for agency.}
\label{fig:agency_prompt}
\end{figure*}

\begin{figure*}
    \setlength{\fboxsep}{10pt}
    \fbox{%
    \begin{minipage}{\linewidth}
    \small

    \vspace{8pt}
    \textbf{SETTING DIMENSIONS}
     
    \vspace{4pt}
    These four dimensions capture how the text constructs a sense of place, time, and physical presence.
     
    \vspace{6pt}
    \textbf{6. Concreteness}
     
    How concrete is the language of the text?
     
    A higher score reflects text in which most content could be explained by pointing, demonstrating, or showing (objects, physical states, bodily actions, perceptual qualities). A lower score reflects text in which most content can only be explained using other words (categories, principles, relationships, institutions, ideas). The key distinction is not just whether physical things are present, but how much the text renders them. When a passage enumerates many specific, tangible, pointable things, the cumulative effect creates strong concrete presence.
     
    \begin{itemize}[noitemsep, leftmargin=*]
        \item Score 5: \textit{``the chipped blue mug sat on the edge of the sink, handle cracked from years of use''}
        \item Score 3: \textit{``She had been commuting to the same office for six years. The building was on a corner downtown, glass and steel, indistinguishable from the ones on either side of it.''}
        \item Score 1: \textit{``people often form attachments to everyday objects''}
    \end{itemize}
     
    \vspace{6pt}
    \textbf{7. Temporal Grounding}
     
    How strongly does the text create a sense of being anchored in a particular time?
     
    A higher score reflects text that creates a vivid sense of temporal location. A lower score reflects text that feels as if events could be taking place at any time. Two types contribute to the score: \textit{historical grounding} locates the reader in a specific, unrepeatable moment (a year, an era, a named event); \textit{cyclical grounding} locates the reader within a recurring temporal structure (a season, a time of day). Historical grounding can reach the upper end of the scale even with sparse language; cyclical grounding alone typically reaches a 3.
     
    \begin{itemize}[noitemsep, leftmargin=*]
        \item Score 5: \textit{``It was the summer of 2015 at 9:02am, before the world changed forever.''}
        \item Score 3: \textit{``This unique summer school gives teenagers the opportunity to work with tutors and musicians.''}
        \item Score 1: \textit{``he walked outside.''}
    \end{itemize}
     
    \vspace{6pt}
    \textbf{8. Spatial Grounding}
     
    How strongly does the text create a sense of being anchored in a particular place?
     
    A higher score reflects text that creates a vivid sense of spatial location. A lower score reflects text that feels as if events could be taking place anywhere. Two types contribute to the score: \textit{geographic grounding} locates the reader on the globe (a country, city, named landmark); \textit{proximate grounding} locates the reader in an immediate physical environment (a room, a building, a street). Geographic grounding can reach a 5 without proximate rendering; vividly rendered proximate grounding without geographic grounding caps at a 4.
     
    \begin{itemize}[noitemsep, leftmargin=*]
        \item Score 5: \textit{``In the narrow streets of Naples, the apartment's tiled floors kept cool even in August heat.''}
        \item Score 3: \textit{``She worked at an office in London.'' or ``The kitchen was small, cluttered with dishes.''}
        \item Score 1: \textit{``She decided to quit her job.''}
    \end{itemize}
     
    \vspace{6pt}
    \textbf{9. Sensory}
     
    How central are sensory details to the text?
     
    A higher score reflects text in which one or more senses are central to how the passage is organized, where the sensory experience drives, anchors, or sustains the content. A lower score reflects text where events and ideas are conveyed without meaningful appeal to the senses. What matters is both whether a sense modality is central to the passage, and whether it is vividly described.
     
    \begin{itemize}[noitemsep, leftmargin=*]
        \item Score 5: \textit{``the bread was still warm, its crust crackling under her fingers, the whole kitchen thick with the smell of it''}
        \item Score 3: \textit{``The bread was still warm when she pulled it from the oven. She set it on the rack and went to wash her hands.''}
        \item Score 1: \textit{``she made breakfast which consisted of bread and eggs.''}
    \end{itemize}

     \vspace{8pt}
    \textbf{Scoring Scale:} 1 = Not at all \quad 2 = Slightly \quad 3 = Moderately \quad 4 = Considerably \quad 5 = Extremely
    
    \end{minipage}
    }
    \caption{Annotation prompt for setting.}
    \label{fig:setting_prompt}
\end{figure*}
\clearpage
\subsection{LLM Annotation Prompt: Event Relations}
\label{app:llm_prompt_event}
 
Fig.~\ref{fig:event_prompt} displays the prompt used to elicit temporal ordering and causal relation annotations from Claude Sonnet~4.6 and Gemma4-31B at scale. The prompt was held constant across both models and across LLM validation and large-scale annotation.
 
\vspace{6pt}
\begin{figure*}[t]
\setlength{\fboxsep}{10pt}
\fbox{%
\begin{minipage}{\linewidth}
    \small
     
    \textbf{TEMPORAL ORDERING}
     
    \vspace{4pt}
    Which event started first in the narrative timeline?
     
    \begin{itemize}[noitemsep, leftmargin=*]
        \item \textbf{span1\_first} --- SPAN1 started before SPAN2
        \item \textbf{span2\_first} --- SPAN2 started before SPAN1
        \item \textbf{simultaneous} --- Both events genuinely share a start point and are distinct events
        \item \textbf{same\_event} --- The two spans refer to the same event from different angles
        \item \textbf{too\_hard\_to\_tell} --- The start times could feasibly be in either order, or the events occur in completely unrelated temporal frames
    \end{itemize}
     
    If either span is not an event, use \textbf{not\_applicable}.
     
    \vspace{4pt}
    \textit{Inferring order.} Use explicit connectives (``before,'' ``after,'' ``then,'' ``when,'' ``while,'' ``as''), backward-looking cues (``preceded,'' ``followed,'' ``prior to''), and logical necessity. A cause precedes its effect, a question is asked before it is answered, a proposal precedes an agreement. When one event logically presupposes the other, infer the ordering.
     
    \vspace{4pt}
    \textit{Reporting and retrospective verbs.} Story-world events precede the speech acts or memory acts that describe them, even when the reporting verb appears first in the text. \textit{``The coach expressed postgame that losing the player had impacted the offense''} $\to$ span2\_first. Memory verbs (\textit{recalls, remembers, reflects}) similarly occur after the events they describe.
     
    \vspace{4pt}
    \textit{Simultaneous vs.\ sequential.} Use simultaneous only when neither event initiates the other and the text marks coincident starts (``just as X, Y''). When ``as'' or ``while'' appears, ask whether both events genuinely begin at the same instant ($\to$ simultaneous) or whether one was already underway when the other began ($\to$ span1\_first or span2\_first). Process/result pairs and physical chains are sequenced even when compressed.
     
    Example: \textit{``While touring in Germany, he released a record.''} $\to$ span1\_first. Touring was already underway when the release happened; ``while'' marks an ongoing background activity, not a coincident start.
     
    \vspace{4pt}
    \textit{Same event.} Use same\_event when both spans describe the same occurrence and neither adds new information the other omits. Continuation verbs (\textit{adding, continuing, noting}) extend a prior speech act $\to$ use span1\_first.
     
    \vspace{4pt}
    \textit{Too hard to tell.} Use when: (a) two events describe different aspects of the same episode with no logical ordering between them; (b) the events involve different unrelated actors with no shared context; or (c) the ordering requires world knowledge inference the text does not support. If you find yourself inferring order from narrative position or plausible sequence rather than explicit textual evidence, use too\_hard\_to\_tell. Uncertainty is not a failure.
     
    \vspace{10pt}
    \textbf{CAUSAL RELATION}
     
    \vspace{4pt}
    Given a confirmed temporal ordering, how are the two events causally related?
     
    \begin{itemize}[noitemsep, leftmargin=*]
        \item \textbf{direct\_cause} --- E1 is sufficient to produce E2; given E1, E2 was bound to happen without any intervening decision or action.
        \begin{itemize}[noitemsep, leftmargin=*]
            \item Physical chain: \textit{``She dropped her phone. The screen cracked.''}
            \item Involuntary reaction: \textit{``He read the message. His stomach dropped.''}
        \end{itemize}
        \item \textbf{enables} --- E1 creates a necessary precondition for E2: if E1 had not happened, E2 could not have happened in the same way. The dependency must be traceable in the text; mere co-occurrence is not enough. Enablement can run in either direction.
        \item \textbf{not\_related} --- No traceable causal dependency. Events that are merely co-present in the narrative without a direct conditional link.
    \end{itemize}
     
    \vspace{4pt}
    \textit{Story-world vs.\ discourse-world.} Story-world events (actions, occurrences, mental states) and discourse events (\textit{said, reported, wrote, announced}) operate at different levels and cannot causally relate to each other. A fire does not cause a spokesperson's statement; the statement is an independent communicative act. When one span is a reporting verb and the other is a story-world event, use \textbf{not\_related}. Two discourse events can causally relate to each other.
     
    \vspace{4pt}
    \textit{Distinguishing tip.} Was E2 bound to happen given E1 alone? $\to$ direct\_cause. Did E1 create a clear necessary precondition? $\to$ enables. Otherwise $\to$ not\_related.
     
\end{minipage}%
}
\caption{Annotation prompt for event relations.}
\label{fig:event_prompt}
\end{figure*}

\end{document}

%% file: commands.tex
\definecolor{maria-color}{HTML}{7881F2}

\usepackage{xspace}
\definecolor{todo-color}{HTML}{ff3232}

\usepackage{xparse}
\NewDocumentCommand{\todocite}{g}{%
  {\color{todo-color}[CITE\IfNoValueF{#1}{: #1}]}\xspace%
}

\usepackage{booktabs}
\usepackage{array}

\usepackage{float}
\usepackage{subcaption}
\usepackage{graphicx}

\usepackage{fontawesome5}

\definecolor{applications-color}{HTML}{e6d4fa}
\definecolor{theory-color}{HTML}{cff5f5}
\definecolor{proposal-color}{HTML}{d4fae6}
\definecolor{4-color}{HTML}{f9d4fa}
\definecolor{5-color}{HTML}{fae6d4}
\definecolor{warn-color}{HTML}{f9d4fa}

\definecolor{lightblue}{RGB}{212, 235, 255}
\definecolor{orange}{RGB}{255, 105, 0}
\definecolor{lightgreen}{RGB}{177, 231, 171}
\definecolor{lightyellow}{RGB}{255, 255, 148}